\documentclass[10pt,journal,compsoc]{IEEEtran}

\usepackage{graphicx}
\usepackage{amsmath}
\usepackage{amssymb}
\usepackage{bm}
\usepackage{booktabs}
\usepackage{multirow}
\usepackage{xcolor}
\usepackage{cleveref}
\newcommand{\myparagraph}[1]{\textbf{#1}}
\usepackage{subfig}

\ifCLASSOPTIONcompsoc
  \usepackage[nocompress]{cite}
\else
  \usepackage{cite}
\fi
\ifCLASSINFOpdf
\else
\fi
\begin{document}
%
\title{Sparsity-guided Network Design for Frame Interpolation}
%
%
%
%

\author{Tianyu~Ding*,~\IEEEmembership{Member,~IEEE,}
        Luming~Liang*, Zhihui~Zhu, Tianyi~Chen,
        and~Ilya~Zharkov
\IEEEcompsocitemizethanks{\IEEEcompsocthanksitem *Equal contribution.
\IEEEcompsocthanksitem T. Ding, L. Liang, T. Chen, and I. Zharkov are with Microsoft, Seattle, USA (e-mails: tianyuding@microsoft.com; lulian@microsoft.com; tianyi.chen@microsoft.com; zharkov@microsoft.com)
\IEEEcompsocthanksitem Z. Zhu is with the Department of Computer Science and Engineering, the Ohio State University, Columbus, USA. E-mail: zhu.3440@osu.edu}
\thanks{}}

%
%

\markboth{}%
{}
%



\IEEEtitleabstractindextext{%
\begin{abstract}
DNN-based frame interpolation, which generates intermediate frames from two consecutive frames, is often dependent on model architectures with a large number of features, preventing their deployment on systems with limited resources, such as mobile devices. We present a compression-driven network design for frame interpolation that leverages model pruning through sparsity-inducing optimization to greatly reduce the model size while attaining higher performance. Concretely, we begin by compressing the recently proposed AdaCoF model and demonstrating that a 10$\times$ compressed AdaCoF performs similarly to its original counterpart, where different strategies for using layerwise sparsity information as a guide are comprehensively investigated under a variety of hyperparameter settings. We then enhance this compressed model by introducing a multi-resolution warping module, which improves visual consistency with multi-level details. As a result, we achieve a considerable performance gain with  a quarter of the size of the original AdaCoF. In addition, our model performs favorably against other state-of-the-art approaches on a wide variety of datasets. We note that the suggested compression-driven framework is generic and can be easily transferred to other DNN-based frame interpolation algorithms. The source code is available at \url{https://github.com/tding1/CDFI}.
\end{abstract}

}

\maketitle

\IEEEdisplaynontitleabstractindextext

%
\IEEEpeerreviewmaketitle

\ifCLASSOPTIONcompsoc
\IEEEraisesectionheading{\section{Introduction}\label{sec:introduction}}
\else
\section{Introduction}
\label{sec:introduction}
\fi

%
%
%
%
\IEEEPARstart{V}{ideo} frame interpolation is a low-level computer vision task that involves creating interim (non-existent) frames between actual frames in a sequence to greatly improve the temporal resolution. It plays an important role in many applications, including frame rate up-conversion~\cite{bao2018high,geng2022rstt}, slow-motion generation~\cite{jiang2018super}, and novel view synthesis~\cite{flynn2016deepstereo,zhou2016view}. Though fundamental, the problem is challenging in that the complex motion, occlusion and feature variation in real world videos are difficult to estimate and 
predict in a transparent way. 

\begin{figure}[]
    \centering
    \includegraphics[width=3in]{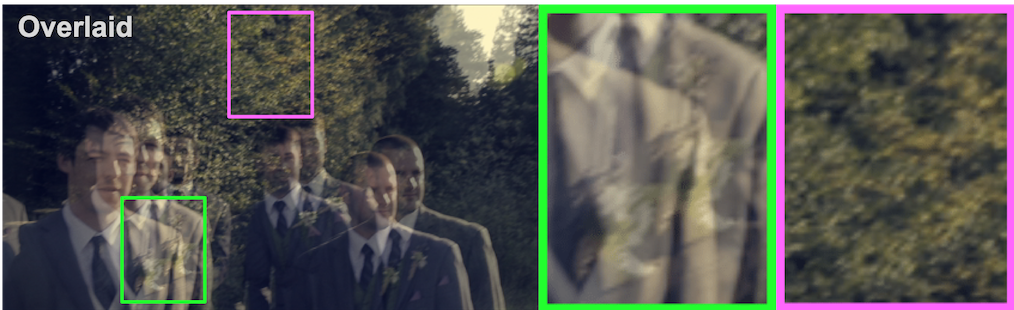}\\
    \includegraphics[width=3in]{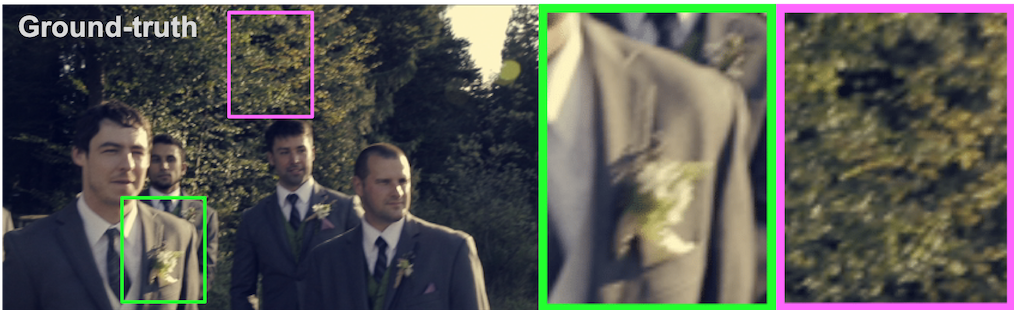}\\
     \includegraphics[width=3in]{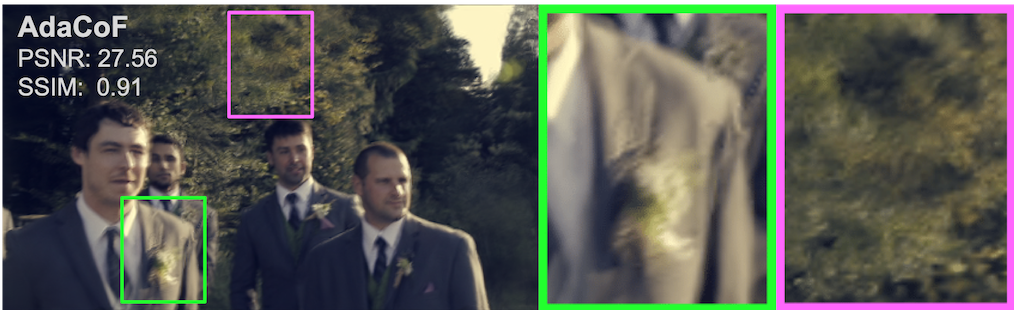}\\
     \includegraphics[width=3in]{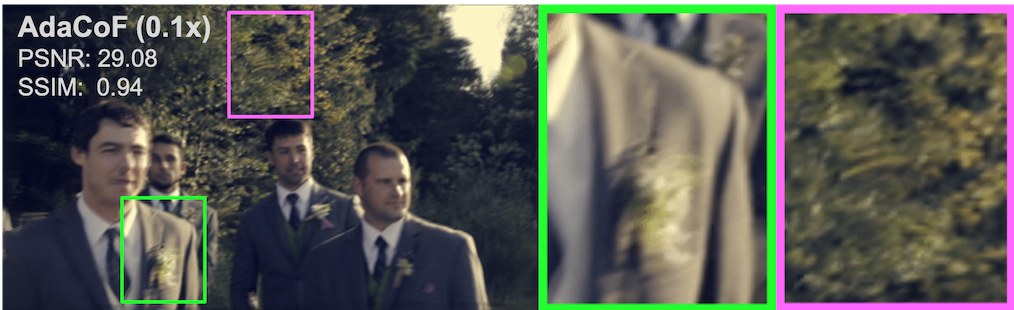}\\
     \includegraphics[width=3in]{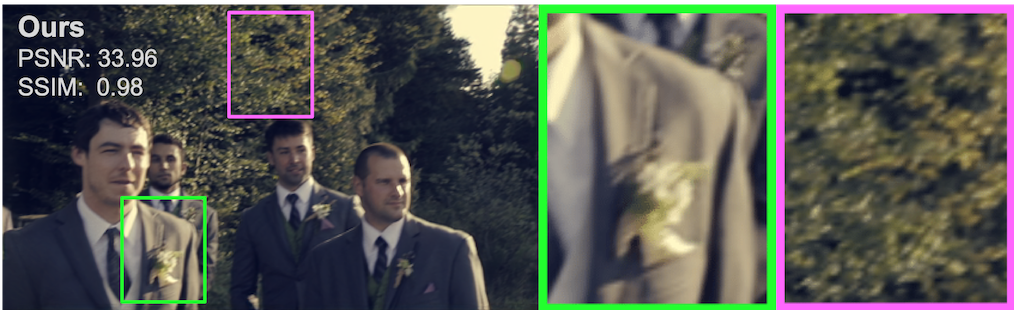}
    \caption{\small\textbf{A challenging example consists of large motion, severe occlusion and non-stationary finer details.} Top to bottom: the overlaid two inputs, the ground-truth middle frame, the frame generated by AdaCoF~\cite{lee2020adacof},
    the frame generated by the 10$\times$ compressed AdaCoF, and the frame generated by our method. The compressed AdaCoF even outperforms the full one here.}
    \label{fig:example}
\end{figure}

Recently, a large number of researches have been conducted in this field, especially those based on deep neural networks (DNN) for their promising outcomes in motion estimation~\cite{dosovitskiy2015flownet,ilg2017flownet,sun2018pwc,weinzaepfel2013deepflow}, occlusion reasoning~\cite{bao2019depth,jiang2018super,peleg2019net} and image synthesis~\cite{dosovitskiy2015learning,flynn2016deepstereo, kalantari2016learning,kulkarni2015deep,zhou2016view}. In particular, due to the rapid expansion in optical flow~\cite{baker2011database, werlberger2011optical}, many approaches either utilize an off-the-shelf flow model~\cite{bao2019depth,niklaus2018context,niklaus2020softmax,xu2019quadratic,hu2022many,niklaus2022splatting} or estimate their own task-specific flow~\cite{jiang2018super,liu2017video,xue2019video,yuan2019zoom,park2020bmbc, huang2020rife, danier2022st,danier2022enhancing,zhang2} as pixel-level motion interpolation guidance. However, integrating a pre-trained flow model makes the entire architecture cumbersome, and task-oriented flow alone is still insufficient in handling sophisticated occlusion and blur with only pixel-level input. Kernel-based approaches~\cite{niklaus2017video,niklaus2017videosepcov,peleg2019net}, on the other hand, synthesize intermediate frames by performing convolution operations over local patches surrounding each output pixel. Nevertheless, it is incapable of handling large motions beyond the kernel size and it typically suffers from high computational cost. There are also hybrid methods as~\cite{bao2019depth,bao2019memc} that combine the advantages of flow-based and kernel-based methods, but the networks are significantly heavier and thus limiting their applications.

We have noticed an increasing trend toward designing more intricate and heavy DNN-based models for interpolating video frames. The majority of the methods proposed in recent years~\cite{bao2019depth,bao2019memc,cheng2020video,choi2020channel,jiang2018super,lee2020adacof,niklaus2017videosepcov,xu2019quadratic} entail training and inference on DNN models with over 20 million parameters. 
The hybrid MEMC-Net~\cite{bao2019memc}, for example, has about 70 million parameters and takes up about 280 megabytes when stored in 32-bit floating point. Large models are typically difficult to train and inefficient during inference. Furthermore, they are unlikely to be deployed on mobile devices, which severely limits their application possibilities. Meanwhile, other work~\cite{chi2020all,liu2017video,xue2019video,yuan2019zoom} concentrate on simple and light-weight video interpolation methods. However, they either perform less competitively on benchmark datasets or are restricted to a specific architecture that is difficult to adapt.

\begin{figure}[]
\centering
\includegraphics[width=3in]{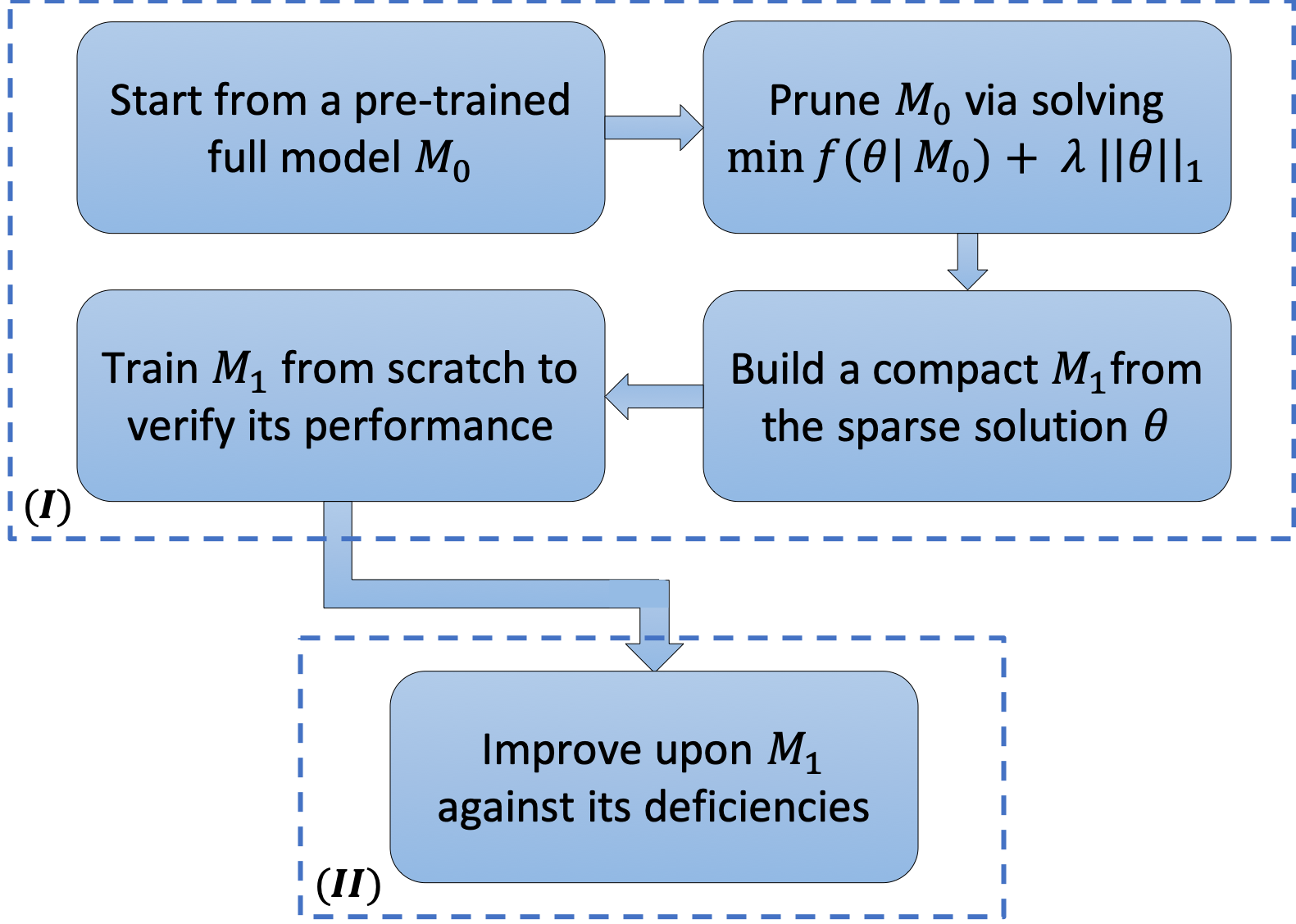}
\caption{\small\textbf{Pipeline of the framework.} Stage (I): compression of the baseline; Stage (II): improvements upon the compression.}
\label{fig:compress}
\vspace{-.1in}
\end{figure}

In this paper, we propose a sparsity-guided network design for video interpolation that exploits model compression~\cite{bucilu2006model,cheng2017survey,zhu2017prune}. 
Concretely, we compress the recently proposed \mbox{AdaCoF}~\cite{lee2020adacof} via fine-grained pruning~\cite{zhu2017prune} based on sparsity-inducing optimization~\cite{chen2020orthant}, and demonstrate that a 10$\times$ compressed \mbox{AdaCoF} is still capable of maintaining the same benchmark performance as before, indicating that the original model contains a substantial amount of redundancy. \emph{The compression provides us with two direct benefits: (i) it facilitates a thorough understanding of the model architecture, which in turn inspires an efficient design; and (ii) the resulting compact model leaves more room for further enhancements that could potentially lead to a breakthrough in performance.} Observing that \mbox{AdaCoF} 
can handle large motion but is incapable of dealing with occlusion or preserving finer details, we enhance the compact model by incorporating a multi-resolution warping module that utilizes a feature pyramid representation of the input frames
to aid in image synthesis. Consequently, our final model outperforms \mbox{AdaCoF} on three benchmark datasets by a substantial margin ($>1$ dB of PSNR on Middlebury~\cite{baker2011database}), despite being a quarter the size of its initial version. Note that it is often challenging to implement the same enhancements on the original heavy model. Experiments show that our model also performs favorably against other state-of-the-arts.

In summary, we present a {compression-driven} paradigm for video interpolation, in which we reflect on over-parameterization. First, we compress \mbox{AdaCoF} and obtain a compact model with comparable performance, then we improve upon it (see the pipeline in Figure~\ref{fig:compress}). This approach yields superior performance and can be easily transferred to other DNN-based frame interpolation algorithms.

A preliminary version of this work was published in
CVPR’21~\cite{ding2021cdfi}. Compared to the conference version, this paper makes the following additional technical contributions:
\begin{itemize}
    \item We illustrate in detail how we utilize optimization-based fine-grained pruning on a baseline model, i.e., AdaCoF, under a variety of hyperparameter settings.
    \item We propose different strategies for compressing the baseline model given the layer-by-layer sparsity information obtained during the pruning process.
    \item We conduct extensive experiments on benchmark datasets to justify the effectiveness of making improvements upon the compact model.
\end{itemize}

We remark that one of the primary goals of the paper is to elaborate on the method of compressing neural networks by sparsity-inducing optimization using comprehensive experiments, which is omitted from~\cite{ding2021cdfi}, and how it may be leveraged to improve the performance of frame interpolation. The remaining sections are organized as follows. \Cref{sec:work} reviews related works. \Cref{sec:approach} explains in detail the sparsity-guided compression and the subsequent improvements on AdaCoF. We conduct extensive experiments in \Cref{sec:exp} and conclude the paper in \Cref{sec:con}.

\section{Related Work}\label{sec:work}

\subsection{Video Frame Interpolation}


Conventional video frame interpolation is modeled as an image sequence problem, such as the path-based~\cite{mahajan2009moving} and phase-based approach~\cite{meyer2018phasenet,meyer2015phase}. Unfortunately, due to their inability to effectively estimate the path~(flow) or represent high-frequency components, these techniques are less successful in complicated scenarios.

Convolutional neural network (CNN) has recently proved its success in understanding temporal motion~\cite{dosovitskiy2015flownet,ilg2017flownet,raket2012motion,sun2018pwc,weinzaepfel2013deepflow,werlberger2011optical} by predicting optical flow, resulting in flow-based motion interpolation methods. \cite{long2016learning}~trains a deep CNN to directly synthesize the intermediate frame. \cite{liu2017video} estimates the flow by sampling the 3D spatio-temporal neighborhood of every output pixel. \cite{kong2022ifrnet, huang2020rife,danier2022st} refine the estimation of intermediate flows in order to capture large and complex motions. \cite{jiang2018super,park2020bmbc,xue2019video,yuan2019zoom,zhang2}~utilize bi-directional flows to warp frames and additional modules to address occlusion. \cite{niklaus2018context,niklaus2020softmax,niklaus2022splatting} integrate an off-the-shelf flow model~\cite{sun2018pwc} into the network, while~\cite{niklaus2020softmax,niklaus2022splatting} suggest a differentiable forward mapping as opposed to the backward mapping employed by many other approaches. 
Instead of assuming uniform motion with linear interpolation,  quadratic~\cite{liu2020enhanced,xu2019quadratic} and cubic~\cite{chi2020all} non-liner models are proposed to estimate complex motions. Similarly, \cite{liu2022atca} proposes an arc trajectory based model that learns motion prior from two successive frames.

One major drawback of the flow-based methods is that only pixel-wise information is used for interpolation. In contrast, kernel-based approaches offer to construct the image by convolving across local patches in close proximity to each output pixel. For instance, \cite{niklaus2017videosepcov}~estimates  spatially-adaptive 2D convolution kernels and \cite{niklaus2017video}~improves its efficiency by employing pairs of 1D kernels for all output pixels simultaneously. \cite{bao2019depth,bao2019memc}~integrate both optical flow and local kernels; specifically~\cite{bao2019depth} detects the occlusion with depth information. Nevertheless, these approaches rely solely on local kernels and are incapable of handling large motion outside the rectangular kernel region.


Inspiring by the flexible spatial sampling locations of deformable convolution (DConv)~\cite{dai2017deformable,zhu2019deformable}, \cite{lee2020adacof}~introduces the AdaCoF model, which synthesizes each output pixel using a spatially-adaptive separable DConv. \cite{shi2020video}~generalizes it by allowing sampling in the entire spatial-temporal space. \cite{cheng2020video}~is similar to AdaCoF except that it approximates 2D kernels using 1D separable kernels. \cite{cheng2020multiple}~extends~\cite{cheng2020video} to construct the intermediate frame at an arbitrary time step. Besides, a recent work based on DConv \cite{danier2022enhancing} uses 3D CNN at multi-scale for estimating complex motions. This paper is also based on AdaCoF; however, unlike the prior work, for the first time we investigate the over-parameterization issue in existing DNN-based techniques and demonstrate that a much smaller model works comparably well through compression. Moreover, by addressing its shortcomings upon the compression, one can easily build a model (still small) that significantly outperforms the original one. This compression-driven network design is 
transferable to other DNN-based frame interpolation techniques.

\begin{figure}[]
{\small \hspace{-.2in} Ground-truth\hspace{.2in} AdaCoF~\cite{lee2020adacof}\hspace{.3in} Ours}
    \centering    \includegraphics[width=0.36\textwidth]{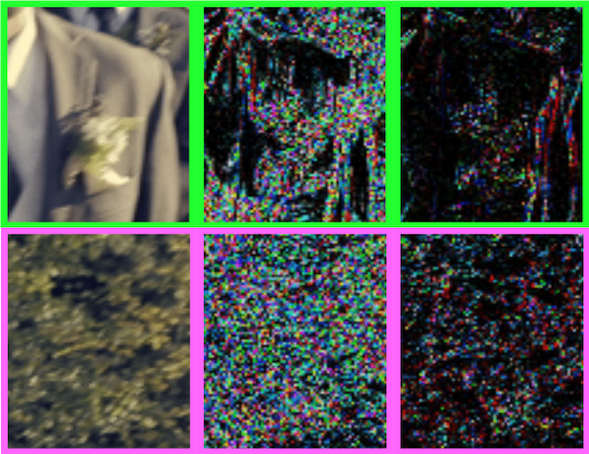}
    \caption{\small\textbf{Visualization of the difference between the interpolation and the ground-truth image.}}
    \label{fig:feat}
    \vspace{-.1in}
\end{figure}

\subsection{Pruning-based Model Compression}

Model compression~\cite{bucilu2006model,cheng2017survey} is especially critical to DNN models, which are known to incur large storage and computation costs. In general, model compression may be categorized into several types: pruning~\cite{zhu2017prune}, quantization~\cite{polino2018model}, knowledge distillation~\cite{hinton2015distilling} and AutoML~\cite{he2018amc}. In this paper, we employ the pruning strategy due to its ease of use, which seeks to induce sparse connections. There are numerous hybrid pruning approaches that directly aiming for model deployment. For example, \cite{chen2015compressing}~proposes a HashNets architecture to group connection weights with a low cost hash function; \cite{han2015deep} presents a deep compression pipeline consists of pruning, quantization and Huffman coding; \cite{ullrich2017soft} suggests a regularization-based approach for soft weight-sharing. However, they may be unnecessary for our purposes of searching and constructing an architecture \emph{after the compression}. In fact, compression serves an entirely different role in our work, working as a tool for gaining a deeper understanding of the underlying architecture and allowing for additional enhancements. We therefore focus on optimization-based sparsity-inducing pruning techniques~\cite{lebedev2016fast,li2016pruning,wen2016learning,zhou2016less} that incorporate sparsity-constrained training, such as with $\ell_0$ or $\ell_1$ regularizers. Specifically, we employ a simple three-step pipeline (see Stage (I) in Figure~\ref{fig:compress}) that is most similar to~\cite{chen2020neural,han2015learning} and consists of: (i) training with $\ell_1$-norm sparsity constraint; (ii) reformulating a small dense network according to the sparse structures identified in each layer; and (iii) retraining the small network to verify its performance. We will see shortly (Sec. \ref{sec:compression}) that both its implementation and test are simple.

\section{The Proposed Approach}\label{sec:approach}

Given two consecutive frames $I_0$ and $I_1$ in a video sequence, the goal of frame interpolation is to synthesize an intermediate frame $I_t$, where $t\in(0,1)$ is an arbitrary temporal location. A typical practice is $t = 0.5$, that is synthesizing the middle frame between $I_0$ and $I_1$. We now introduce the proposed framework with AdaCoF~\cite{lee2020adacof} as an instance.

\begin{figure*}[!ht]
    \centering
    \includegraphics[width=0.99\textwidth]{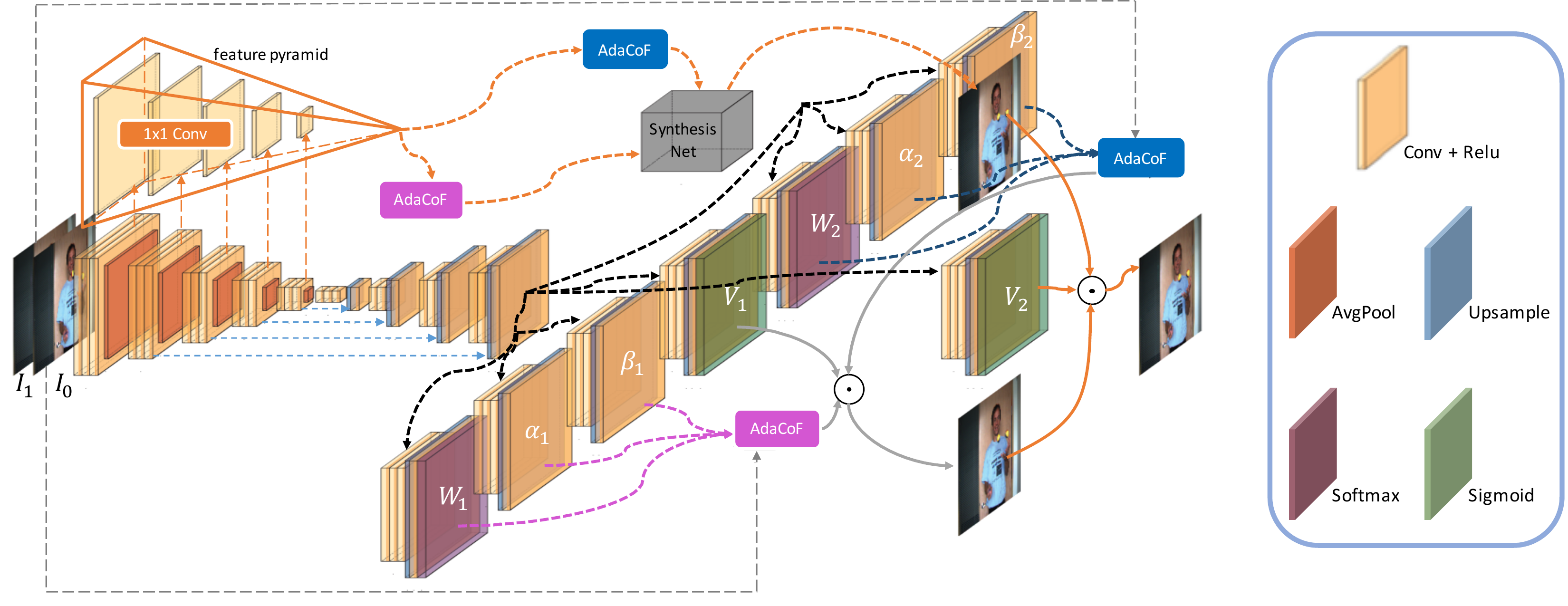}
    \caption{\small\textbf{Illustration of our architecture design based on the compressed AdaCoF~\cite{lee2020adacof}.} The lower part (AdaCoF) consists of a U-Net, a group of sub-networks for estimating two sets of $\{W_i,\alpha_i,\beta_i\}$ in~\eqref{eq:adacof} that correspond to backward/forward warping, and an occlusion mask $V_1$ for synthesizing one candidate intermediate frame $I_{0.5}^{(1)}$. The upper part (our design) extracts a feature pyramid of the input frames via 1-by-1 convolutions from the encoder of the U-Net, then the multi-scale features are warped by AdaCoF operation of learned backward/forward parameters, which are fed to a synthesis network to generate another candidate intermediate frame $I_{0.5}^{(2)}$. Note that the pink and blue AdaCoF modules are associated with $\{W_1,\alpha_1,\beta_1\}$ and $\{W_2,\alpha_2,\beta_2\}$, respectively. The network generates the final result by blending $I_{0.5}^{(1)}$ and $I_{0.5}^{(2)}$ via an extra occlusion mask $V_2$.}
    \label{fig:model}
    \vspace{-.1in}
\end{figure*}

\subsection{Motivation}
 
To illustrate AdaCoF, we first introduce one of its major components, a spatially-adaptive separable DConv operation for synthesizing one image (denoted by $I_{\text{out}}$) from another one (denoted by $I_{\text{in}}$). In order to generate  $I_{\text{out}}$ from $I_{\text{in}}$,  the input image $I_{\text{in}}$ is padded so that $I_{\text{out}}$ retains the original shape of $I_{\text{in}}$.  For each pixel $(i,j)$ in $I_{\text{out}}$, AdaCoF computes $I_{\text{out}}(i, j)$  by convolving a deformable patch surrounding the reference pixel $(i,j)$ in $I_{\text{in}}$:
 \begin{align}\label{eq:adacof}
\sum_{k=0}^{F-1}\sum_{l=0}^{F-1}
W_{i,j}^{(k,l)}I_{\text{in}}\big(i+dk+\alpha_{i,j}^{(k,l)},j+dl+\beta_{i,j}^{(k,l)}\big),
\end{align}
where $F$ is the deformable kernel size, $W_{i,j}^{(k,l)}$ is the $(k,l)$-th kernel weight in synthesizing $I_{\text{out}}(i, j)$, $\vec\Delta:=\big(\alpha_{i,j}^{(k,l)},\beta_{i,j}^{(k,l)}\big)$ is the offset vector of the $(k,l)$-th sampling point associated with $I_{\text{in}}(i, j)$, and $d\in\{0,1,2,\cdots\}$ is the dilation parameter that helps to explore a wider area. Note that $F$ and $d$ have pre-determined values. For synthesizing each output pixel in $I_{\text{out}}$, a total of $F^2$ points are sampled in $I_{\text{in}}$. With the offset vector $\vec\Delta$, the $F^2$ sample points are not confined to a rectangular region centered on the reference point. On the other hand, unlike the classic DConv, AdaCoF uses various kernel weights across multiple reference pixels $(i,j)$, indicated by $W_{i,j}^{(k,l)}$ in~\eqref{eq:adacof}; hence the attribute ``separable''~\cite{niklaus2017videosepcov}. 

Since the parameters $\{W_{i,j}^{(k,l)}, \alpha_{i,j}^{(k,l)},\beta_{i,j}^{(k,l)}\}$ are computed individually for each output pixel, AdaCoF is flexible in handling large and complex motion; however, it cannot deal with severe occlusion and non-stationary finer details, as shown in Figure~\ref{fig:example}. We further visualize the difference between the interpolation and the ground-truth in Figure~\ref{fig:feat}. AdaCoF is insufficient for maintaining contextual information because the interpolation is simply produced by blending the two warped frames with a sigmoid mask ($V_1$), as demonstrated in Figure~\ref{fig:model}. A natural question to ask is that if direct improvements are possible. In fact, we find the architecture design of the AdaCoF model is relatively cumbersome, especially the encoder-decoder part. For instance, six $512\times512\times 3\times 3$ convolutional layers are utilized in the center, which is an entire heuristic since it is unclear whether or not this design is adequate for the interpolation task. When  $F=5, d=1$, the original AdaCoF model has 21.8 million parameters and takes 83.4 megabytes if stored with PyTorch. Typically, training and validating such a large model takes a considerable amount of time, preventing direct improvements upon it. To better understand the architecture and improve its performance, we propose the following compression-driven approach.

\subsection{First Stage: Compression of the Baseline}\label{sec:compression}

\subsubsection{General Methodology}\label{subsec:genral-method}

As the first stage in our approach, we compress the baseline model by leveraging the fine-grained model pruning~\cite{zhu2017prune} via sparsity-inducing optimization~\cite{chen2018fast}. Specifically, given a pre-trained full model $M_0$, we begin by re-training (fine-tuning) its weights~$\theta$ by applying an $\ell_1$ norm sparsity regularizer and solving the following optimization problem:
\begin{align}\label{eq:sparse_prob}
 \min_{\theta}\ f(\theta | M_0) + \lambda \|\theta\|_1,
\end{align}
where $f(\cdot)$ denotes the training objective for our task (see Sec.~\ref{sec:training} for details) and $\lambda>0$ is the regularization constant. It is known that, when $\lambda$ is chosen suitably, the formulation~\eqref{eq:sparse_prob} promotes a sparse solution, allowing one to easily identify  the connections between neurons that correspond to non-zero weights. In order to solve~\eqref{eq:sparse_prob}, we utilize a recently proposed orthant-based stochastic method termed OBProx-SG~\cite{chen2020orthant}, which offers an efficient mechanism for encouraging sparsity and less performance regression than existing solvers. Solving the $\ell_1$-regularized problem~\eqref{eq:sparse_prob} results in a fine-grained pruning, as zeros are promoted in an unstructured manner. Note that it is also possible to impose group sparsity constraints~\cite{chen2020half,chen2021only,lebedev2016fast,zhou2016less}, such as mixed $\ell_1/\ell_2$, to prune the kernel weights in a group-wise manner. We only adopt the $\ell_1$ 
constraint in the presentation due to its ease of use.

\begin{figure*}[]
\vspace{-.1in}
    \centering
    \subfloat[][With all Proximal Steps]{\includegraphics[width=0.42\textwidth]{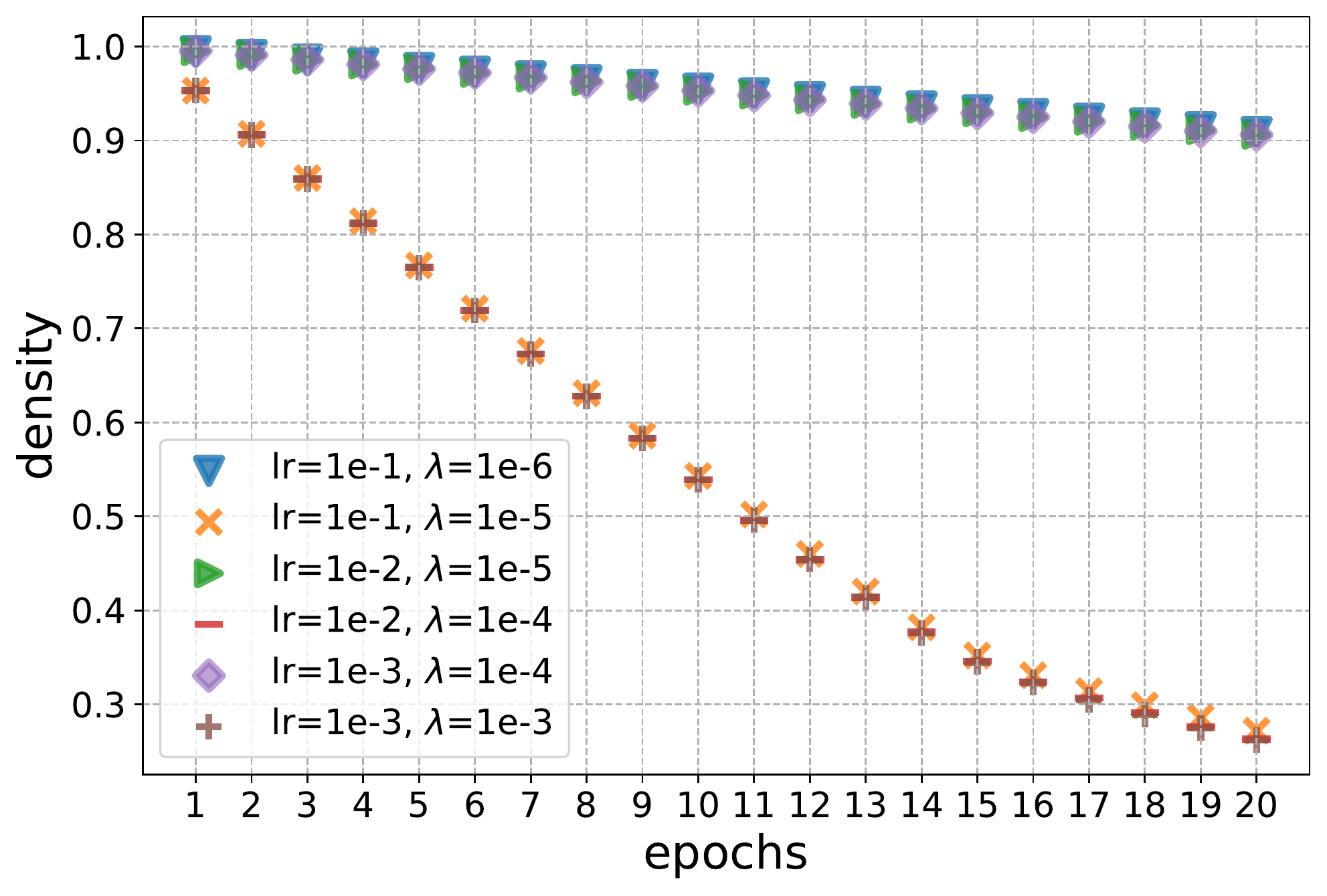}\label{fig:prune-20-a}}\qquad 
    \subfloat[][With half Proximal Steps and half Orthant Steps]{\includegraphics[width=0.42\textwidth]{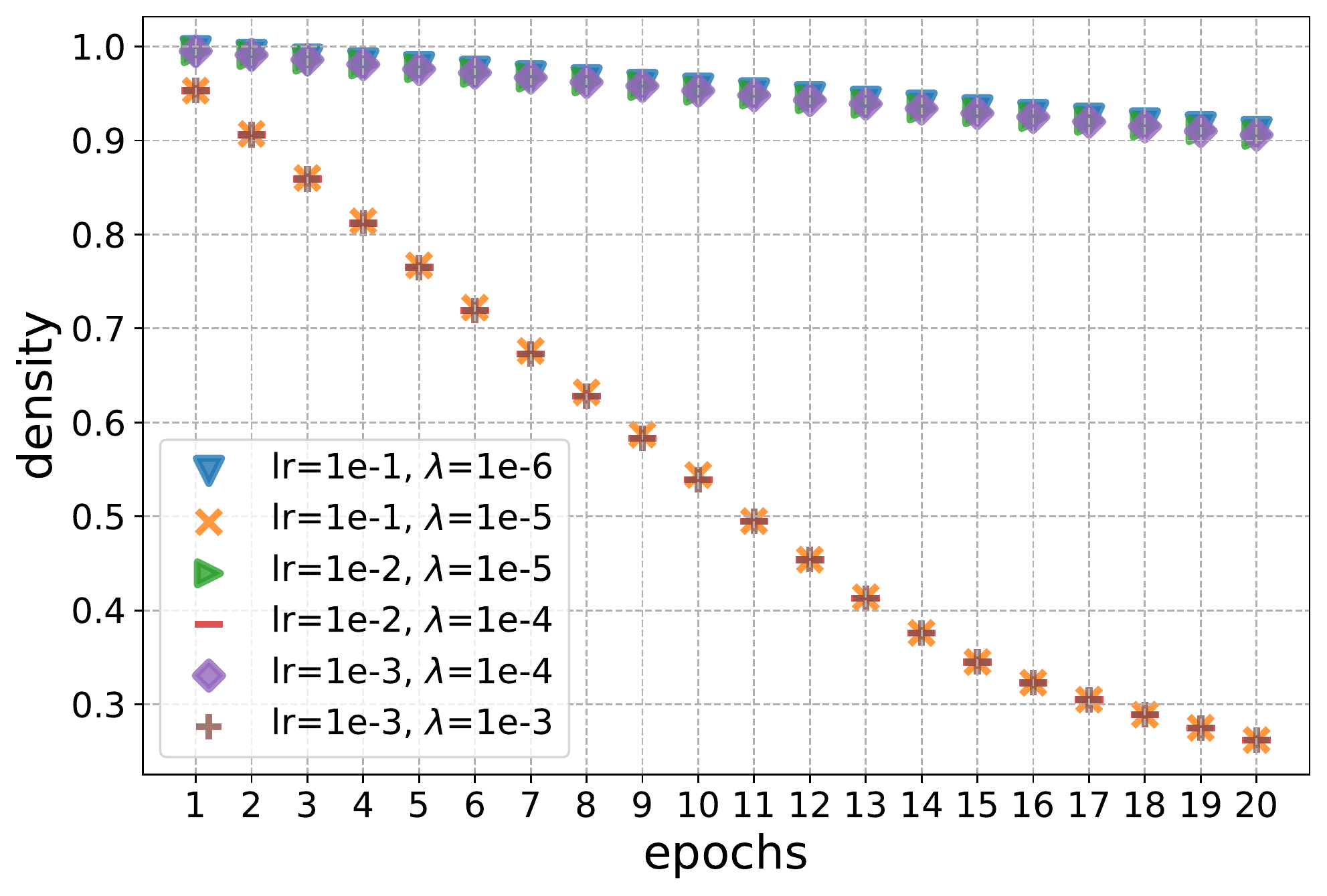}\label{fig:prune-20-b}}
    \caption{\small\textbf{Plot of density of AdaCoF at the first 20 epochs when optimizing problem~\eqref{eq:sparse_prob} with OBProx-SG.}}
    \label{fig:prune-20}
    \vspace{-.1in}
\end{figure*}

\begin{figure*}[]
    \centering
    \subfloat[][With all Proximal Steps]{\includegraphics[width=0.42\textwidth]{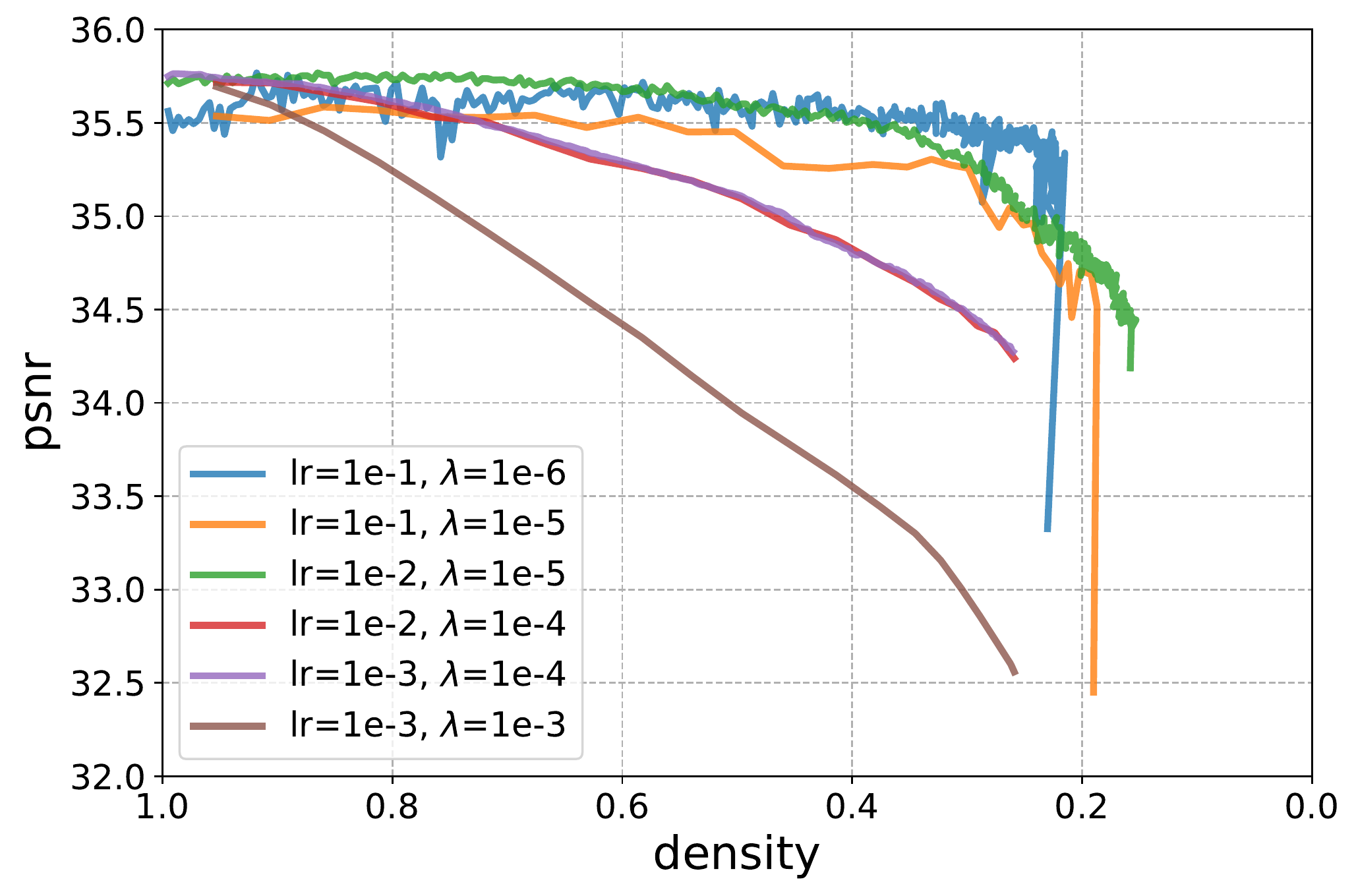}\label{fig:prune-a}}\qquad 
    \subfloat[][With half Proximal Steps and half Orthant Steps]{\includegraphics[width=0.42\textwidth]{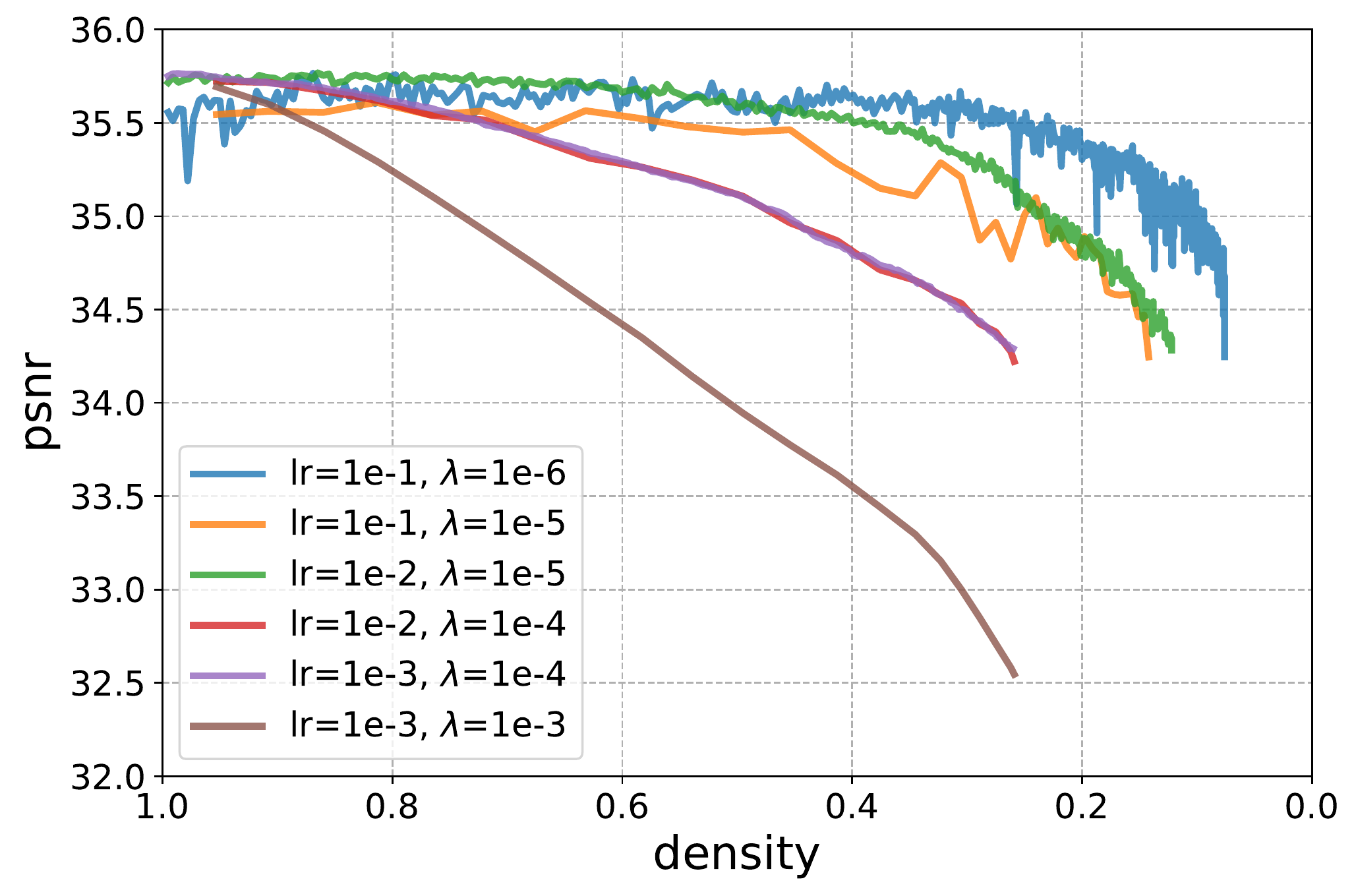}\label{fig:prune-b}}
    \caption{\small\textbf{Plot of PSNR (evaluated on Middlebury) against the density of AdaCoF when optimizing problem~\eqref{eq:sparse_prob} with OBProx-SG.}}
    \label{fig:prune}
    \vspace{-.1in}
\end{figure*}

After obtaining a sparse solution $\hat \theta$, distinct from~\cite{han2015learning} that acts directly on a sparse network, we re-design a small dense network $M_1$ depending on the sparsity determined at each layer. We first illustrate the concept with convolutional layers, however the same practice may also be applied to fully connected layers. Given the $l$-th convolutional layer consisting of $K_l=C_l^{\text{in}}\times C_l^{\text{out}}\times q\times q$ parameters (denoted as $\hat\theta_l$), where $C_l^{\text{in}}$ is the number of input channels, $C_l^{\text{out}}$ is the number of output channels, $q\times q$ is the kernel size, then the sparsity $s_l$ and density ratio $d_l$ of this layer are defined as 
\begin{align}
 s_l:= \big(\text{\# of zeros in } \hat\theta_l\big)/{K_l}\quad\text{and}\quad
 d_l:= 1-s_l.
\end{align}
Inspired by~\cite{chen2020neural}, we use $d_l$ as the \emph{compression ratio} and reconstruct the layer with shape\footnote{The formulation of~\eqref{eq:dl} is slightly different than that proposed in~\cite{ding2021cdfi} but is more reasonable, which is used throughout the paper.}
\begin{align}\label{eq:dl}
\left\lceil \sqrt{d_l}\cdot C_l^{\text{in}}\right\rceil \times \left\lceil \sqrt{d_l}\cdot C_l^{\text{out}}\right\rceil \times q \times q
\end{align}
so that the current number of parameters is roughly $d_l$ times fewer than it was previously. The main intuition is that the density ratio $d_l$ reflects the least amount of necessary information that must be encoded in that layer without significantly impacting performance according to the sparsity-inducing optimization results. Similarly, for a fully connected layer with $H_{\text{in}}$ input features and $H_{\text{out}}$ output features, we can reconstruct one linear layer with shape $\lceil\sqrt{d_l}\cdot H_{\text{in}}\rceil \times\lceil\sqrt{d_l}\cdot H_{\text{out}}\rceil $. Note that we exclude the bias term in both convolutional and linear layers when counting parameters, as the density ratio is used as an inexact guide and is sufficient for the reconstruction of a compact model.

Let $\widetilde C_{l}^{\text{out}}$ and $\widetilde C_{l+1}^{\text{in}}$ denote the newly updated number of output channels at the $l$-th layer and the number of input channels at the $(l+1)$-th layer. One outstanding question is that one cannot simply reformulate each layer in the manner described before, such as~\eqref{eq:dl} for convolutional layers, because a valid neural network architecture also requires $\widetilde C_{l}^{\text{out}}\equiv\widetilde C_{l+1}^{\text{in}}$.  In other words, it is not guaranteed that $\left\lceil \sqrt{d_l}\cdot C_l^{\text{out}}\right\rceil\equiv \left\lceil \sqrt{d_{l+1}}\cdot C_{l+1}^{\text{in}}\right\rceil$. In light of the fact that the density ratio in each layer is only used as a rough guide for network design, we propose the following two strategies, which are referred as ``-min" and ``-max" for abbreviation:
\begin{itemize}
    \item ``-min": the minimum of $\lceil \sqrt{d_l}\cdot C_l^{\text{out}}\rceil$ and $ \lceil \sqrt{d_{l+1}}\cdot C_{l+1}^{\text{in}}\rceil$ is chosen such that 
    {\small
    \begin{align}
        \widetilde C_{l}^{\text{out}}=\widetilde C_{l+1}^{\text{in}}=\min\{\lceil \sqrt{d_l}\cdot C_l^{\text{out}}\rceil,  \lceil \sqrt{d_{l+1}}\cdot C_{l+1}^{\text{in}}\rceil \}    \end{align}
        }
    \item ``-max": the maximum of $\lceil \sqrt{d_l}\cdot C_l^{\text{out}}\rceil$ and $ \lceil \sqrt{d_{l+1}}\cdot C_{l+1}^{\text{in}}\rceil$ is chosen such that 
    {\small
    \begin{align}
        \widetilde C_{l}^{\text{out}}=\widetilde C_{l+1}^{\text{in}}=\max\{\lceil \sqrt{d_l}\cdot C_l^{\text{out}}\rceil,  \lceil \sqrt{d_{l+1}}\cdot C_{l+1}^{\text{in}}\rceil \}    \end{align}
        }
\end{itemize}
We are then able to recreate a valid model architecture utilizing either of the two ways. Indeed, the above strategies are applicable to anyplace in the network accordingly where the number of input/output features or channels do not match. An example is the shortcut connections between the encoder and decoder of the U-Net, as shown in~\Cref{fig:model}.

Finally, to evaluate the performance of the compressed model $M_1$, we train it from scratch (without the $\ell_1$ constraint). Due to its reduced size, the compact model typically requires substantially less time to train than that of the full model $M_0$. The entire compression pipeline is depicted in Stage (I) of Figure~\ref{fig:compress}. We remark that a pre-trained $M_0$ is not essential for the purpose of compression since problem~\eqref{eq:sparse_prob} is sufficient for a one-shot training/pruning, but $M_0$ allows us to ensure the compressed model 
performs competitively.

\begin{table*}[]
\caption{\small\textbf{
The statistics of AdaCoF and the compressed versions using "-min" and "-max" strategies.
}}
\label{tab:compress}
\vspace{-.1in}
\begin{center}
\begin{tabular}{cc|ccccc|ccccc}
\toprule               & \begin{tabular}[c]{@{}c@{}} AdaCoF\\ $(F=5,d=1)$
\end{tabular} & \begin{tabular}[c]{@{}c@{}} lr:\\ $\lambda$: \end{tabular} & \begin{tabular}[c]{@{}c@{}} $10^{-2}$\\ $10^{-4}$ \end{tabular}  & \begin{tabular}[c]{@{}c@{}} $10^{-1}$\\ $10^{-5}$ \end{tabular} & \begin{tabular}[c]{@{}c@{}} $10^{-2}$\\ $10^{-5}$ \end{tabular} &
\begin{tabular}[c]{@{}c@{}} $10^{-1}$\\ $10^{-6}$ \end{tabular} &
\begin{tabular}[c]{@{}c@{}} lr:\\ $\lambda$: \end{tabular}&
\begin{tabular}[c]{@{}c@{}} $10^{-2}$\\ $10^{-4}$ \end{tabular}  & \begin{tabular}[c]{@{}c@{}} $10^{-1}$\\ $10^{-5}$ \end{tabular} & \begin{tabular}[c]{@{}c@{}} $10^{-2}$\\ $10^{-5}$ \end{tabular} & \begin{tabular}[c]{@{}c@{}} $10^{-1}$\\ $10^{-6}$ \end{tabular} \\
\midrule
PSNR                                                           & {35.72}   & \multirow{6}{*}{"-min"}    & 35.22     &  35.27   & 35.15   &  34.24 & \multirow{6}{*}{"-max"} & 35.56 &  35.57 &  35.47  &  34.28 \\
SSIM                                                           & {0.96}    &     & {0.96}     & 0.95  & 0.95  &   0.95 &  & 0.96 & 0.96  & 0.96   &  0.95  \\
Size (MB)                                                      & 83.4   &      &     18.2       &  9.2  &  7.8  &  4.9 & & 26.7 & 14.8 & 12.8  & 7.9 \\
Time (ms)                                                      & 61.6   &      &   49.5    &   43.1     & 42.0  & 38.9 & & 50.2 &  46.5 &  45.9  &  41.3 \\
FLOPS (G)                                                      & 359.2   &     &  235.6        &  183.1   & 169.8  &  142.5 &  &  255.4 &  205.2 & 194.0 &  163.6  \\
 \begin{tabular}[c]{@{}c@{}} Params (M)\end{tabular} & 21.84    &     & 4.64      &  2.38   &  2.02  &  1.25 & &  6.83 &  3.78 & 3.27 &  2.04 \\
 \bottomrule
\end{tabular}
\end{center}
\vspace{-.1in}
\end{table*}


\begin{figure}[]
    \centering    \includegraphics[width=0.42\textwidth]{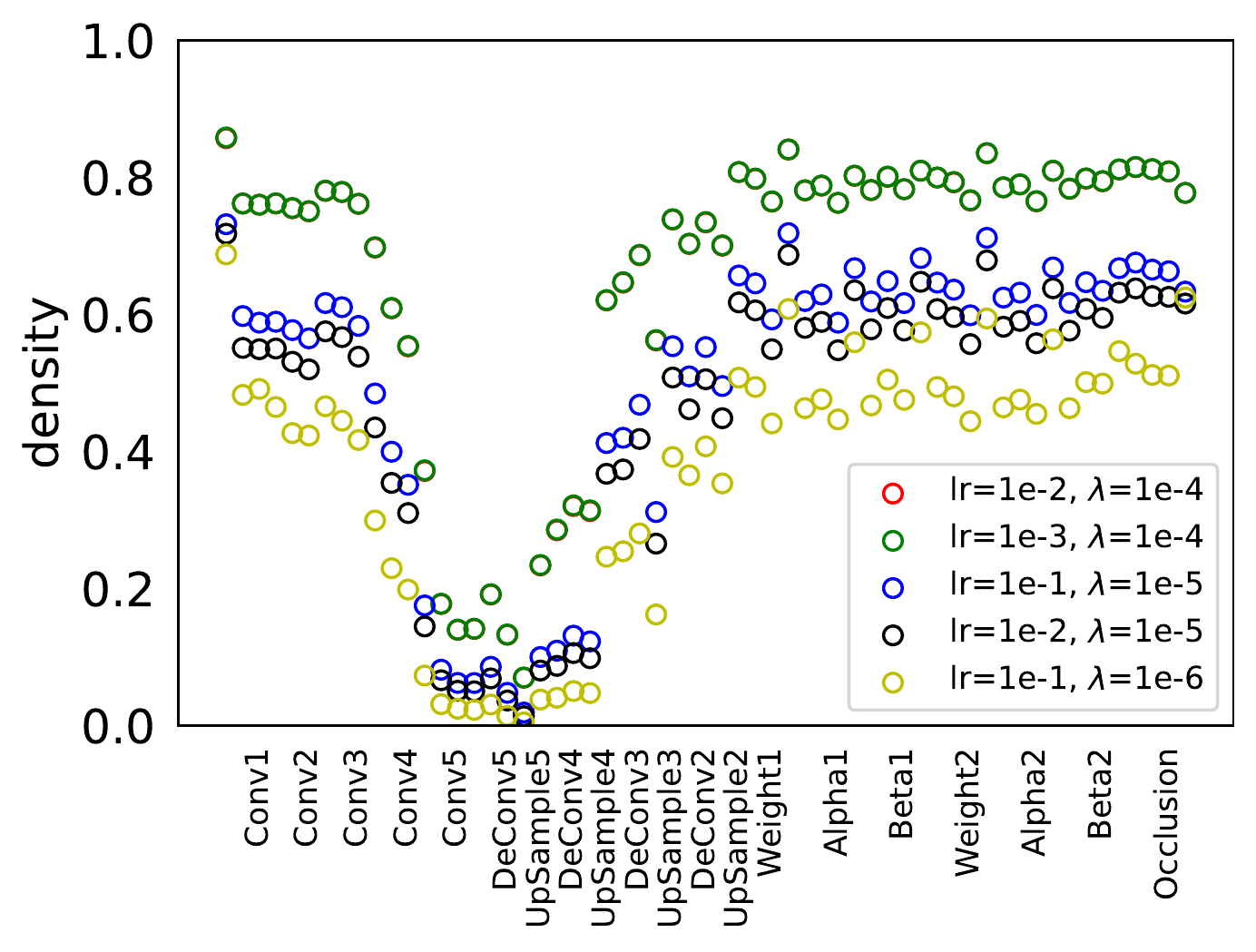}
    \caption{\small\textbf{Density distributions for each layer under various hyperparameter settings.}}
    \label{fig:prune-layer}
    \vspace{-.1in}
\end{figure}

\subsubsection{Compression of AdaCoF} 

We now apply the generic compression approach introduced in~\Cref{subsec:genral-method} to AdaCoF~\cite{lee2020adacof}, utilizing the pre-trained model provided by the authors. To tackle problem~\eqref{eq:sparse_prob}, we employ the orthant-based stochastic method OBProx-SG~\cite{chen2020orthant}, which is shown to be superior to other state-of-the-art methods~\cite{xiao2009dual,xiao2014proximal} in finding a sparse optimal solution to the $\ell_1$ problem.  Specifically, it includes so-called Proximal Steps (P-steps) and Orthant Steps (O-steps) during the iterative optimization process, where P-steps produce reasonable solutions but with less sparsity and O-steps are the key to encourage more sparse solutions. In the following, we investigate the performance of OBProx-SG under various hyperparameter settings, such as different learning rate (lr) and regularization constant ($\lambda$), and the manner in which P-steps/O-steps are conducted. During optimization, we only use 1000 video triplets from Vimeo-90K~\cite{xue2019video}.

\myparagraph{The impact of hyperparameters.}  For problem~\eqref{eq:sparse_prob}, the regularization constant $\lambda$ plays a crucial role in controlling the sparsity level of the solution since a large $\lambda$ penalizes more on the $\ell_1$ term, resulting in a sparser solution. In addition, learning rate (lr) is also critical to the optimization result. In fact, in OBProx-SG, the sparsity of the solution is determined by the magnitude of $\text{lr}\cdot \lambda$. In \Cref{fig:prune-20}, we plot the density of AdaCoF across the first 20 epochs when solving~\eqref{eq:sparse_prob} with OBProx-SG. We observe that, for a fixed learning rate, raising $\lambda$ introduces significantly sparser solutions at each iteration. In the mean time, the several $(\text{lr}, \lambda)$ pairs corresponding to a same magnitude of $\text{lr}\cdot\lambda$ result in solutions with similar densities at each iteration. For learning rates and $\lambda$ such that $\text{lr}\cdot\lambda=10^{-6}$ the density declines to below 30 percent within 20 epochs however for the other settings  $\text{lr}\cdot\lambda=10^{-7}$ the density decreases more smoothly. In order to understand how the model performs as the optimization proceeds, we display the PSNR evaluated on the Middlebury dataset~\cite{baker2011database} versus the network density, as demonstrated in Figure~\ref{fig:prune}. One can detect a decline in model performance as sparsity increases. Consider the two extreme instances. When $\text{lr}=10^{-3},\lambda=10^{-3}$, not only does the density decline rapidly (\Cref{fig:prune-20}), but so does the PSNR (\Cref{fig:prune}). In contrast, when $\text{lr}=10^{-1},\lambda=10^{-6}$, even while the density decreases slowly (\Cref{fig:prune-20}), it is able to retain a reasonable level of model performance (\Cref{fig:prune}) at a low sparsity level. In practice, there is a trade-off between a model performance with less regression (a small $\lambda$) and a short training time. Specifically, when $\text{lr}\cdot\lambda=10^{-6}$, it only takes about 20 epochs to achieve a model with a density of less than 30 percent, however when $\text{lr}\cdot\lambda=10^{-7}$, it may take hundreds of epochs to attain a similar density level.

\begin{table*}[!ht]\footnotesize
\caption{\small\textbf{Comprehensive quantitative results of our full models under various settings.}}
\label{tab:quant-new}
\vspace{-.1in}
\begin{center}
\begin{tabular}{ccccccccccccc}
\toprule 
\multirow{2}{*}{\begin{tabular}[c]{@{}c@{}}Compressed  \\ arch. w/ ($\text{lr},\lambda$)\end{tabular}}              &  \multirow{2}{*}{\begin{tabular}[c]{@{}c@{}}Which \\ strategy?\end{tabular}}    &  \multirow{2}{*}{\begin{tabular}[c]{@{}c@{}}Other params. \\ ($F, d$)\end{tabular}}                                     & \multicolumn{3}{c}{Vimeo-90K~\cite{xue2019video}} & \multicolumn{3}{c}{Middlebury~\cite{baker2011database}} & \multicolumn{3}{c}{UCF101-DVF~\cite{liu2017video}} &
\multirow{2}{*}{\begin{tabular}[c]{@{}c@{}}Parameters\\ (million)\end{tabular}} \\ \cmidrule(lr){4-6} \cmidrule(lr){7-9} \cmidrule(lr){10-12}
&  &  & PSNR      & SSIM    & LPIPS   & PSNR    & SSIM    & LPIPS    & PSNR     & SSIM    & LPIPS    &              \\ 
\midrule
\multicolumn{2}{c}{Original AdaCoF~\cite{lee2020adacof}} & $(5,1)$   &  34.35   &    0.956    &  0.019     &  35.72    & 0.959     & 0.019  & 35.16     & \textbf{0.950}    & {0.019}  &  21.84  \\ 
\midrule
$(10^{-2},10^{-4})$     &  "-min"  & $(5,1)$   &  34.71   &    0.960    &  0.011     &  36.56    & 0.963     & {0.008}   & 35.16     & {0.949}    & \textbf{0.015}  &  7.05  \\ 
$(10^{-1},10^{-5})$     &  "-min"  & $(5,1)$   & 34.82    & 0.961       &   0.011    &    36.21  & 0.962    & 0.008     &   35.12   &  0.949    &  \textbf{0.015}  &  4.77  \\ 
$(10^{-2},10^{-5})$     &  "-min"  & $(5,1)$  &  35.02   &    0.962    &  {0.011}     &  36.34  & 0.962    &  {0.008}     & 35.15     & {0.949}   & \textbf{0.015} &  4.41  \\ 
$(10^{-1},10^{-6})$     &  "-min"  & $(5,1)$   &   34.86  & 0.961       &  0.011     &   36.31   & 0.962  & 0.008  &  35.14    &   0.949   &    \textbf{0.015}     & 3.63  \\ 
\midrule
$(10^{-2},10^{-4})$     &  "-max"  &  $(5,1)$  &  34.97   &  0.962        &    \textbf{0.010} &   36.43   &  0.963 &  {0.008}   &   35.14   &{0.949}    &  \textbf{0.015} & 9.25  \\ 
$(10^{-1},10^{-5})$     &  "-max"  &  $(5,1)$   &  34.98   &  0.962      & \textbf{0.010}  &    36.23  & 0.962    &   {0.008}   &   35.14   & {0.949}  & \textbf{0.015} & 6.18  \\ 
$(10^{-2},10^{-5})$     &  "-max"  &  $(5,1)$   &  34.81    &  0.961      &  0.011      &   36.62   & 0.964    &  {0.008}  & 35.15     & {0.949}    & \textbf{0.015} &  5.66  \\ 
$(10^{-1},10^{-6})$     &  "-max"  & $(5,1)$    &  34.93   &  0.962    &  \textbf{0.010}  &  36.66     &  0.964    & {0.008}      & 35.18     & \textbf{0.950}     & \textbf{0.015}  & 4.42 \\ 
\midrule
\multicolumn{2}{c}{Original AdaCoF+~\cite{lee2020adacof}} & $(11,2)$   &  34.56   &    0.959    &  0.018     &  36.09    & 0.962     & 0.017  & 35.16     & \textbf{0.950}     & {0.019} &  22.93  \\ 
\midrule
$(10^{-2},10^{-4})$     &  "-min"  & $(11,2)$    &   35.09  & {0.963}    &     \textbf{0.010}&  37.12  & \textbf{0.967}    &  \textbf{0.007}    & \textbf{35.22}    & \textbf{0.950}     & \textbf{0.015} &  8.09  \\ 
$(10^{-1},10^{-5})$     &  "-min"  & $(11,2)$    &  {35.14}   &  {0.963}    & \textbf{0.010}    &   \textbf{37.25}     & \textbf{0.967}    &   \textbf{0.007}  &  35.20    & \textbf{0.950}    & \textbf{0.015}  &  5.78 \\ 
$(10^{-2},10^{-5})$     &  "-min"  & $(11,2)$    &  35.02    & {0.963}   &  \textbf{0.010}    &  36.98    & {0.966}     &   \textbf{0.007}   &  35.20    & \textbf{0.950}     & \textbf{0.015}   & 5.41 \\ 
$(10^{-1},10^{-6})$     &  "-min"  & $(11,2)$    & 35.07    &  {0.963}    &    \textbf{0.010}   &  37.05    & {0.966}    &  \textbf{0.007}    & {35.21}     & \textbf{0.950}     & \textbf{0.015}  &  4.61 \\ 
\midrule
$(10^{-2},10^{-4})$     &  "-max"  & $(11,2)$    & \textbf{35.22}    &  \textbf{0.964}      &\textbf{0.010}     &  37.17    &  \textbf{0.967}    &  \textbf{0.007}   & {35.21}     & \textbf{0.950}      & \textbf{0.015} & 10.30  \\ 
$(10^{-1},10^{-5})$     &  "-max"  &$(11,2)$    &  35.06   &  {0.963}      &   \textbf{0.010} &  {37.23}    & \textbf{0.967}    & \textbf{0.007}   & {35.21}     &   \textbf{0.950}    & \textbf{0.015}  &  7.20 \\ 
$(10^{-2},10^{-5})$     &  "-max"  &$(11,2)$    & 35.04    & {0.963}      &   \textbf{0.010}  &  37.14    &  \textbf{0.967}    &  \textbf{0.007}  &  35.19    & \textbf{0.950}    & \textbf{0.015} &  6.68 \\ 
$(10^{-1},10^{-6})$     &  "-max"  & $(11,2)$    & 35.09    &  {0.963}     &   \textbf{0.010} &   37.13   & \textbf{0.967}    & \textbf{0.007}   & {35.21}      &  {0.949}    &  \textbf{0.015}  &  5.41 \\ 
\bottomrule
\end{tabular}
\end{center}
\vspace{-.1in}
\end{table*}

\myparagraph{The impact of Proximal Steps and Orthant Steps.} For the above experiments, we additionally study the effect of how P-steps and and O-steps are conducted. \cite{chen2020orthant} has shown that P-steps can generate good solutions but with less sparsity, whereas O-steps are the key to encouraging sparser solutions. It suggests running O-steps before a sufficient number of P-steps. Recall that in~\Cref{fig:prune} we ran the experiments with $\text{lr}\cdot \lambda=10^{-7}$ for hundreds of epochs and with $\text{lr}\cdot \lambda=10^{-6}$ for tens of epochs. \Cref{fig:prune-a} demonstrates the results when we conduct P-steps throughout each epoch for all the experiments, whereas~\Cref{fig:prune-b} show the results when we conduct half P-steps followed by half O-steps. One can see that for $\lambda=10^{-5}$ and $10^{-6}$, the model trained with half O-steps is able to attain significantly greater sparsity while maintaining a reasonable level of performance, hence demonstrating the advantages of O-steps. Despite the slight difference between~\Cref{fig:prune-20-a} and~\Cref{fig:prune-20-b}, which only depicts the first 20 epochs, a similar phenomena also happens.

\myparagraph{Layer-wise density distribution analysis.} We further investigate the density distribution for each layer under various hyperparameter settings in~\Cref{fig:prune-layer}. The results pertain to models trained with an equal number of P-steps and O-steps (see~\Cref{fig:prune-b}).  We are not plotting the results for $\text{lr}=10^{-3},\lambda=10^{-3}$ due to its inferior performance relative to other settings, as shown in~\Cref{fig:prune}. First, we note that, for all the configurations, the three $512\times 512\times 3\times 3$ convolutional layers (labeled as ``DeConv5") in the center of the U-Net are among the most redundant. Specifically, the entire middle part of the U-Net, beginning roughly with ``Conv5" and ending with ``UpSample4", achieves a density ratio of less than 10\%, indicating that more than 90\% of the kernel is mostly useless. This observation confirms our early hypothesis that the original architecture contains significant redundancy. Another observation is that, the settings with $\text{lr}\cdot \lambda=10^{-7}$ almost have consistently lower density levels in each layer than those with $\text{lr}\cdot \lambda=10^{-6}$. Moreover, smaller $\lambda$ (with longer training time) is always advantageous for inducing sparser connections in each layer. Finally, we note that the results of setting $\text{lr}=10^{-2},\lambda=10^{-4}$ and $\text{lr}=10^{-3},\lambda=10^{-4}$ coincide, which is also evident via \Cref{fig:prune} that the red and purple curves are nearly identical.

\myparagraph{Performance of the reconstructed compact model.} We now reformulate different compact networks based on the computed density ratio in each layer corresponding to various $(\text{lr},\lambda)$ settings during optimization~(\Cref{fig:prune-layer}). We train the models from scratch using the entire taining set (51312 video triplets) of Vimeo-90K. In this case, the training is around $5\times$ faster than it was earlier. Then, we compare the before-and-after models in Table~\ref{tab:compress} for different $(\text{lr},\lambda)$ settings with "-min"/"-max" strategies, where PSNR and SSIM are evaluated on the Middlebury dataset, and time and FLOPS are determined for synthesizing a $3\times 1280\times 720$ frame on RTX 6000 Ti GPU. Notice that the "-max" strategy yields better performance than the "-min" strategy at the expense of a bigger model size and higher computational cost. Interestingly, for the setting $\text{lr}=10^{-1},\lambda=10^{-5}$, even though the PSNR falls below 34.2 during the $\ell_1$ optimization~(\Cref{fig:prune-b}), after reformulation and training the PSNR of the compressed model rises to 35.27 (35.57) with the "-min" ("-max") strategy, which is comparable to the original uncompressed AdaCoF but with a much smaller size ($\sim10\times$ for "-min"). Note that although the setting $\text{lr}=10^{-1},\lambda=10^{-6}$ achieves a higher sparsity during the initial $\ell_1$ optimization, the reconstructed networks do not perform well due to their limited model capacities. 

\begin{table*}[!ht]\footnotesize
\caption{\small\textbf{Ablation experiments on the architecture design of our approach.}}
\label{tab:ablation}
\vspace{-.1in}
\begin{center}
\begin{tabular}{lcccccccccc}
\toprule 
\multirow{2}{*}{}                                                                  & \multicolumn{3}{c}{Vimeo-90K~\cite{xue2019video}} & \multicolumn{3}{c}{Middlebury~\cite{baker2011database}} & \multicolumn{3}{c}{UCF101-DVF~\cite{liu2017video}} & \multirow{2}{*}{\begin{tabular}[c]{@{}c@{}}Parameters\\ (million)\end{tabular}} \\ \cmidrule(lr){2-4} \cmidrule(lr){5-7} \cmidrule(lr){8-10}
                                                                                   & PSNR     & SSIM     & LPIPS    & PSNR      & SSIM     & LPIPS    & PSNR      & SSIM     & LPIPS    &                                                                                 \\ \midrule
AdaCoF ($F=5, d=1$)                                                                  & 34.35    & 0.956    & 0.019    & 35.72     & 0.959    & 0.019    & 35.16     & \textbf{0.950}    & 0.019    & 21.84                                                                            \\ 
Compressed AdaCoF ($F=5, d=1$)     &     34.08     &    0.954      &     0.020     &    35.27       &   0.952       &     0.019     &    35.10       &     \textbf{0.950}     &   0.019       & \textbf{2.38}   \\ \midrule 
AdaCoF+ ($F=11, d=2$)        &     34.56     &    0.959      &     0.018     &     36.09      &   0.962       &     0.017     &     35.16      &     \textbf{0.950}     &    0.019      &           22.93                  \\ 
Compressed AdaCoF+ ($F=11, d=2$)    &      34.33    &   0.957       &    0.019      &  35.63         &  0.959        &  0.018        &    35.12       &   \textbf{0.950}       &    0.019      &  2.48  \\  \midrule
\begin{tabular}[c]{@{}c@{}}Ours: FP ($F=5, d=1$)\end{tabular} &    34.59      &    0.962      &    0.011      &    36.12       &  0.960        &    0.008      &    35.05       &       0.949   &  \textbf{0.015}      &   4.66                                                                              \\
Ours: FP + 1x1 Conv ($F=5, d=1$)    &  34.79        &  \textbf{0.963}       &   0.011       &  36.50         & 0.964         &   0.008       &     35.10      &    0.949      &   \textbf{0.015}    &    4.51                                                                             \\ 
Ours: FP + 1x1 Conv ($F=11, d=2$)                              &    34.99      &   \textbf{0.963}       &   0.011       &  36.88         &  0.966        &  \textbf{0.007}      &  35.17         &    \textbf{0.950}    & \textbf{0.015}       & 5.64                                                                                \\ 
Ours: FP + 1x1 Conv + PS ($F=11, d=2$)   & \textbf{35.14}       &    \textbf{0.963}      &   \textbf{0.010}      &  \textbf{37.25}         &  \textbf{0.967}        &   \textbf{0.007}       &   \textbf{35.20}        &  \textbf{0.950}        &  \textbf{0.015}        &       5.78                                                                          \\ \bottomrule
\end{tabular}
\end{center}
\end{table*}

\begin{figure*}[]
\vspace{-.1in}
    \centering
    \includegraphics[width=0.49\textwidth]{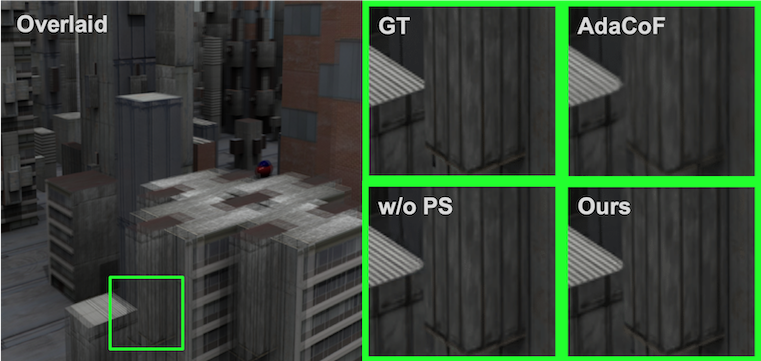} \ \ 
    \includegraphics[width=0.49\textwidth]{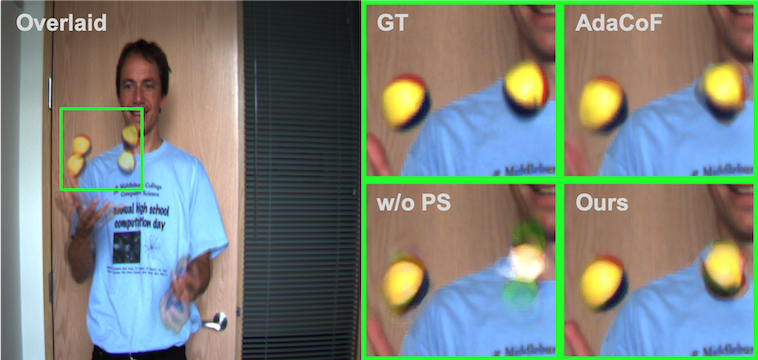}
     \caption{\textbf{Examples of adding the path selection (PS) mechanism in our design.}}
    \label{fig:PS}
\end{figure*}

\subsection{Second Stage: Improve upon the Compression}\label{sec:improve}

In the second stage, we improve upon the compression by addressing its flaws. The point is that the compression allows for additional improvements due to its compactness, which is often impossible when operating directly on the original huge model, e.g., the lengthy training and validation period appears overwhelming in the first place.

Observing that AdaCoF is incapbale of handling severe occlusion and preserving finer features, we design three particular components on top of the compressed AdaCoF: a feature pyramid, an image synthesis network, and a path selection mechanism. Note that improvements are case-by-case, as each baseline model has its own weakness. 

\myparagraph{Feature pyramid.} AdaCoF computes the final interpolation frame by blending the two warped frames through a single sigmoid mask $V_1$ (see Figure~\ref{fig:model}), which is a generalization of using a binary mask to determine the occlusion weights of the two warped frames for each output pixel. We argue that the loss of contextual details in the input frames is unavoidable with only raw pixel information, as it lacks feature space guidance. Instead, we extract from the encoder portion of the U-Net a feature pyramid representation~\cite{niklaus2020softmax} of the input frames. In particular, it includes five feature levels corresponding to the encoder, and for each level, we employ a 1-by-1 convolution to filter the encoder at multi-scale with $4,8,12,16,20$ output features (in descending order by the feature scale). The retrieved multi-scale features are then warped by AdaCoF operation~\eqref{eq:adacof}, which captures motion in the feature space.

\myparagraph{Image synthesis network.} To better utilize the extracted multi-scale features, we synthesize the image using a GridNet~\cite{fourure2017residual} architecture with three rows and six columns, which is also used in~\cite{niklaus2018context,niklaus2020softmax} for its superiority in combining multi-scale information. Specifically, we feed the synthesis network with both the forward- and backward-warped multi-scale feature maps in order to generate a single RGB image that emphasizes contextual features.

\myparagraph{Path selection.} In order to take advantage of both AdaCoF (handling complex motion) and our own components (handling contextual details), we use a path selection mechanism to generate the final interpolation result. As shown in Figure~\ref{fig:model}, one path leads to the output of the original AdaCoF (designated $I_{0.5}^{(1)}$), which is generated by blending two warped input frames with the occlusion mask $V_1$. Another path leads to the output of the synthesis network (designated $I_{0.5}^{(2)}$), which is generated by combining the warped multi-scale feature maps. In the end, we learn another occlusion module $V_2$ to synthesize the final result from $I_{0.5}^{(1)}$ and $I_{0.5}^{(2)}$, and we expect that $I_{0.5}^{(2)}$ to compensate for the lack of contextual information in $I_{0.5}^{(1)}$.

\subsection{Training} \label{sec:training}

With the architecture described above, we train it using AdaMax~\cite{kingma2014adam} with $\beta_1=0.9, \beta_2=0.999$, an initial learning rate of 0.001 that halves every 20 epochs, a mini-batch size of 8, and a maximum of 100 training epochs.

\myparagraph{Objective function.} Given the interpolated frame $I_{\text{out}}$ of our network and its ground truth $I_{\text{gt}}$, we first employ the Charbonnier penalty~\cite{liu2017video} as a surrogate for the $\ell_1$ loss:
\begin{align}
    \mathcal{L}_{\text{Charbon}}=\rho(I_{\text{out}}-I_{\text{gt}})
\end{align}
where $\rho(\bm x)=(\|\bm x\|_2^2+\epsilon^2)^{1/2}$ and $\epsilon$ is set to $0.001$. Next, we follow~\cite{lee2020adacof} and use a perceptual loss with feature $\phi$ extracted from  \texttt{conv4\_3} of the pre-trained VGG16~\cite{simonyan2014very}:
\begin{align}
 \mathcal{L}_{\text{vgg}}=\|\phi(I_{\text{out}})-\phi(I_{\text{gt}})\|_2.
\end{align}
Then, inspired by the implementation of AdaCoF, we apply a total variation loss on the offset vectors to guarantee spatial continuity and smoothness:
\begin{align}
\mathcal{L}_{\text{tv}} = \tau(\bm\alpha_1)+\tau(\bm\alpha_2)+\tau(\bm\beta_1)+\tau(\bm\beta_2)
\end{align}
where $\tau(I)=\sum_{i,j}\rho(I_{i,j+1}-I_{i,j})+\rho(I_{i+1,j}-I_{i,j})$, and $\bm\alpha_1,\bm\alpha_2,\bm\beta_1,\bm\beta_2$ are the offsets modules computed within our network. Lastly, we formulate our final loss function as
\begin{align}
\mathcal{L}= \mathcal{L}_{\text{Charbon}}+ \lambda_{\text{vgg}}\mathcal{L}_{\text{vgg}} + \lambda_{\text{tv}}\mathcal{L}_{\text{tv}}
\end{align}
where we set $\lambda_{\text{vgg}}=0.005,\lambda_{\text{tv}}=0.01$ in the experiments.

\myparagraph{Training dataset.} We use the Vimeo-90K dataset~\cite{xue2019video} for training, which contains 51312/3782 video triplets of size 256$\times 448$ for training/validation. We further augment the data by randomly flipping them horizontally and vertically as well as perturbing the temporal order.

\begin{table*}[!ht]\footnotesize
\caption{\small\textbf{Quantitative comparisons with state-of-the-art methods.} The results of methods marked with $^\dagger$ are cloned from~\cite{niklaus2020softmax}.}
\label{tab:quant}
\vspace{-.1in}
\begin{center}
\begin{tabular}{lccccccccccc}
\toprule 
\multirow{2}{*}{}              & \multirow{2}{*}{\begin{tabular}[c]{@{}c@{}}Training\\ dataset\end{tabular}}                                                     & \multicolumn{3}{c}{Vimeo-90K~\cite{xue2019video}} & \multicolumn{3}{c}{Middlebury~\cite{baker2011database}} & \multicolumn{3}{c}{UCF101-DVF~\cite{liu2017video}} &
\multirow{2}{*}{\begin{tabular}[c]{@{}c@{}}Parameters\\ (million)\end{tabular}} \\ \cmidrule(lr){3-5} \cmidrule(lr){6-8} \cmidrule(lr){9-11}
&  & PSNR      & SSIM    & LPIPS   & PSNR    & SSIM    & LPIPS    & PSNR     & SSIM    & LPIPS    &              \\ 
\midrule
$^\dagger$SepConv - $\mathcal{L}_1$~\cite{niklaus2017videosepcov}     & proprietary & 33.80    & 0.956    & 0.027    & 35.73     & 0.959    & 0.017    & 34.79     & 0.947    & 0.029    & 21.6                                                                            \\ 
$^\dagger$SepConv - $\mathcal{L}_F$~\cite{niklaus2017videosepcov}         & proprietary &   33.45       &    0.951      &    0.019      &    35.03       &  0.954        &  0.013        &   34.69        &  0.945        &  0.024        &      21.6                                                                           \\ 
$^\dagger$CtxSyn - $\mathcal{L}_{Lap}$~\cite{niklaus2018context}      & proprietary &    34.39      &     0.961     &     0.024     &     36.93      &    0.964      &    0.016      &    34.62       &   0.949       &   0.031       &     --                                                                            \\ 
$^\dagger$CtxSyn - $\mathcal{L}_{F}$~\cite{niklaus2018context} &   proprietary       &    33.76      &    0.955      &    0.017       &   35.95       &   0.959       &   0.013        &      34.01    &      0.941    &  0.024    &        --             \\
$^\dagger$SoftSplat - $\mathcal{L}_{Lap}$~\cite{niklaus2020softmax}    &   Vimeo-90K         &     \color{red}\textbf{36.10}     &    \color{red}\textbf{0.970}      &    0.021       &   \color{red}\textbf{38.42}       &  \color{red}\textbf{0.971}        &   0.016        &      \color{red}\textbf{35.39}    &      \color{red}\textbf{0.952}    &      0.033             &     --     \\ 
$^\dagger$SoftSplat - $\mathcal{L}_{F}$~\cite{niklaus2020softmax}          &     Vimeo-90K       &   \color{blue}\textbf{35.48}       &    {0.964}      &    \color{blue}\textbf{0.013}       &   \color{blue}\textbf{37.55}       &  0.965        &   \color{blue}\textbf{0.008}        &  35.10        &  0.948    &  0.022   &  --  \\ 
$^\dagger$DAIN~\cite{bao2019depth}   &    Vimeo-90K        &    34.70      &  {0.964}        &      0.022     &     36.70     &     0.965     &     0.017      &        35.00  &    \color{blue}\textbf{0.950}      &    0.028  &       24.0       \\ 
AdaCoF~\cite{lee2020adacof}       &     Vimeo-90K       &    34.35    & 0.956    & 0.019    & 35.72     & 0.959    & 0.019    & 35.16     & \color{blue}\textbf{0.950}    & {0.019}    & 21.8                           \\ 
AdaCoF+~\cite{lee2020adacof}       &     Vimeo-90K       &    34.56     &    0.959      &     0.018     &     36.09      &   0.962       &     0.017     &     35.16      &     \color{blue}\textbf{0.950}     &    {0.019}     &           22.9             \\
EDSC - $\mathcal{L}_{C}$~\cite{cheng2020multiple}       &     Vimeo-90K       &    34.86      &     {0.962}     &     0.016      &   36.76     &   {0.966}    &    0.014    &      35.17    &  \color{blue}\textbf{0.950}         & {0.019}             &   8.9   \\ 
EDSC - $\mathcal{L}_{F}$~\cite{cheng2020multiple}        &     Vimeo-90K       &    34.57      &    0.958      &    \color{red}\textbf{0.010}       &   36.48       &      {0.963}    &     \color{red}\textbf{0.007}      &   35.04       &   {0.948}      &   \color{red}\textbf{0.015}       &   8.9    \\
BMBC~\cite{park2020bmbc}       &         Vimeo-90K             &   35.06       &    {0.964}      &    0.015      &  36.79       &     {0.965}    &    0.015      &   35.16        &      \color{blue}\textbf{0.950}    &   {0.019}       &    11.0  \\
CAIN~\cite{choi2020channel}       &         Vimeo-90K             &   34.65       &     0.959     &     0.020     &   35.11   &    0.951      &   0.019       &      34.98     &     \color{blue}\textbf{0.950}     &     0.021     & 42.8  \\
DRVI~\cite{wu2021drvi}          &        Vimeo-90K                      &   {35.14}       &    \color{blue}\textbf{0.965}      &    \color{blue}\textbf{0.013}     &  36.74         &  0.964      &   0.011      &   35.13        &  \color{blue}\textbf{0.950}        & \color{blue}\textbf{0.018}        &       \color{red}\textbf{1.3}         \\
RRIN~\cite{li2020video}           &        Vimeo-90K                      &   {35.21}       &    {0.964}      &   0.015     &  35.47        &  0.958       &   0.014      &   34.93       &  0.949        &   {0.019}     &       19.19       \\
$L^2BEC^2$~\cite{zhang2}           &        Vimeo-90K                      &   34.55      &   0.959      &   0.018    &  35.84         &  0.961       &   0.017     &   35.18       &  \color{blue}\textbf{0.950}        &  {0.019}     &       7.1        \\
Ours           &        Vimeo-90K                      &   {35.14}       &    {0.963}      &    \color{red}\textbf{0.010}      &  {37.25}         &  \color{blue}\textbf{0.967}        &   \color{red}\textbf{0.007}       &   \color{blue}\textbf{35.20}        &  \color{blue}\textbf{0.950}        &  \color{red}\textbf{0.015}        &       \color{blue}\textbf{5.8}         \\
\bottomrule
\end{tabular}
\end{center}
\vspace{-.1in}
\end{table*}

\myparagraph{Evaluation.} In addition to the Vimeo-90K validation set, we evaluate the model on the well-known Middlebury dataset~\cite{baker2011database} and UCF101~\cite{liu2017video,soomro2012ucf101}. The metrics we use are PSNR, SSIM~\cite{wang2004image} and LPIPS~\cite{zhang2018unreasonable}. Note that higher PSNR and SSIM values indicate better performance, whereas lower LPIPS values suggest better results.

\myparagraph{Performance of the proposed full model.} Given the sparsity-guided compression~(\Cref{sec:compression}), the improvements~(\Cref{sec:improve}), and the training/evaluation protocol~(\Cref{sec:training}), we compare the full models with all the combinations of compressed baselines, "-min"/"-max" strategies, and $F,d$ (parameters in~\eqref{eq:adacof}) choices. The comprehensive quantitative results are summarized in~\Cref{tab:quant-new}. First, we observe that the proposed sparsity-guided full models perform much better than their AdaCoF counterparts, even with a significantly smaller model size. Second, while having a little larger model, the "plus" versions with $F=11, d=2$ are much superior to those with $F=5,d=1$. Also, despite the fact that the "-max" strategy usually delivers higher performance than the "-min" one, as we observed in~\Cref{tab:compress}, the improvement is not that significant in comparison to the increase in model size. Particularly, we find that the model with a compressed baseline of setting $\text{lr}=10^{-1},\lambda=10^{-5}$, "-min" strategy, and $F=11,d=2$ achieves the overall best quantitative results with only a quarter of the size (5.78M) of the AdaCoF+ (22.93M) while obtaining $>1.1$ dB of PSNR on Middlebury. In the rest of the paper, we refer to this version as "ours" for convenience.



\section{More Experiments} \label{sec:exp}

\subsection{Ablation Study}

We analyze three components in our proposed method: model compression, feature pyramid, and path selection. 

\myparagraph{Model compression.} As described in~\Cref{sec:compression}, we compress the baseline model to eliminate a substantial mount of redundancy, which also facilitates training and inference. In Table~\ref{tab:ablation}, we compare the performance of AdaCoF and its compressed counterpart. It shows that a 10$\times$ compressed model does not sacrifice much when evaluated on the three benchmark datasets, indicating the redundancy in AdaCoF and the necessity of the compression stage.

\myparagraph{Feature pyramid.} In order to better capture the contextual details, we incorporate the feature pyramid (FP) module, followed by warping operations and an image synthesis network, into the compressed AdaCoF. We isolate its effect by training a network that simply outputs the synthesized image without a path selection mechanism. It turns out that using merely the FP module (see ``Ours - FP'',~\Cref{tab:ablation}) improves  PSNR, SSIM and LPIPS significantly on the Vimeo-90K and Middlebury datasets. Note that it substantially improves LPIPS across all the three benchmark datasets. Moreover, filtering the multi-scale feature maps with 1-by-1 convolutions leads to better PSNR and SSIM as well as a slightly smaller model size.

\begin{figure*}[!ht]
{\footnotesize \hspace{0.85in}Ground-truth\hspace{0.9in} Overlaid \hspace{0.12in}AdaCoF+~\cite{lee2020adacof}\hspace{0.04in}BMBC~\cite{park2020bmbc}\hspace{0.1in}CAIN~\cite{choi2020channel}\hspace{0.07in}EDSC-$\mathcal{L}_C$~\cite{cheng2020multiple}\hspace{0.0in} EDSC-$\mathcal{L}_F$~\cite{cheng2020multiple}\hspace{0.1in} \textbf{Ours}  } \\
    \centering
    \includegraphics[width=0.99\textwidth]{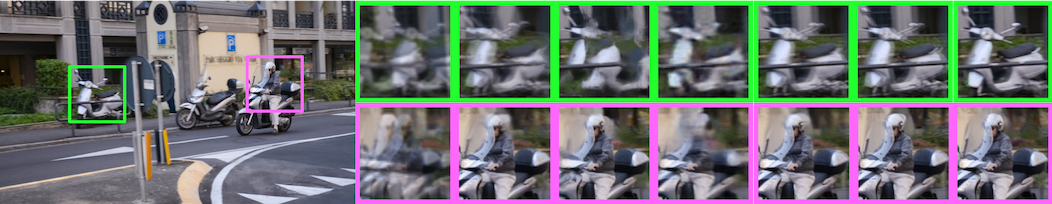}\\
    \includegraphics[width=0.99\textwidth]{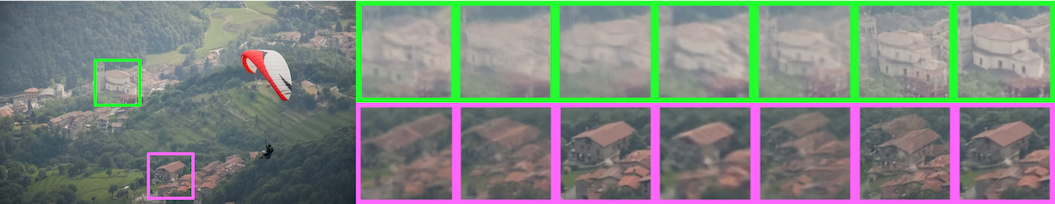}\\
    \includegraphics[width=0.99\textwidth]{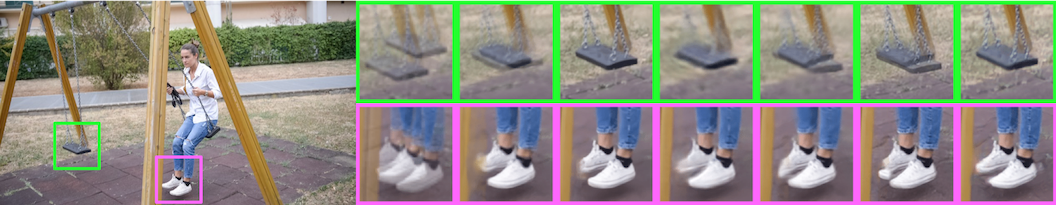}
    \caption{\small\textbf{Visual comparisons on the DAVIS 2016 dataset~\cite{perazzi2016}.} Our sparsity-guided method not only outperforms the baseline model AdaCoF but also surpasses many other methods in handling large motion, occlusion and fine details.}
    \label{fig:visual}
    \vspace{-.1in}
\end{figure*}

\myparagraph{Path selection.} Although by adding only the FP module (and \mbox{1-by-1} convolutions) we can achieve promising quantitative results as shown in~\Cref{tab:ablation}, it does not take advantage of the capability of AdaCoF to handle complex motion, which can be incorporated into our design with the proposed path selection (PS) mechanism. The left example in~\Cref{fig:PS} shows that when there are simply fluctuations in fine detail in the input frames, adding PS or not does not significantly affect our interpolation performance because the FP module can synthesize details (also note the output of AdaCoF is blurry due to the loss of information). On the other hand, with only the FP module, it is difficult for the model to correctly capture the motion of the right ball in the right example, which involves the large motion of two balls. In contrast, our final model, which has a PS mechanism, is able to handle large motion very well (even sharper on the edges of the balls compared to AdaCoF). Importantly, our approach preserves the finger shape (see bottom-left corner), whereas  AdaCoF completely misses this detail. In conclusion, our full model with FP and PS is capable of handling both fine details and big motion, and produces considerable quantitative improvements.

\subsection{Quantitative Evaluation}

We compare our sparsity-guided approach (the compressed AdaCoF with setting $\text{lr}=10^{-1},\lambda=10^{-5}$, "-min" strategy, and $F=11,d=2$) to various state-of-the-art DNN methods in~\Cref{tab:quant}. Since SepCov~\cite{niklaus2017videosepcov}, CtxSyn~\cite{niklaus2018context} and SoftSplat~\cite{niklaus2020softmax} are not open source, we directly copy their numerical results as well as DAIN's~\cite{bao2019depth} from~\cite{niklaus2020softmax}. For the rest of the methods, we evaluate their pre-trained models on the three datasets. 
First note that our approach performs favorably against other methods in terms of SSIM and LPIPS. For PSNR, the proposed method surpasses most competitors with the exception of SoftSplat~\cite{niklaus2020softmax}. Moreover, our model is considerably smaller than those of our competitors. We note that there have been some lightweight frame interpolation models in the past, such as DVF~\cite{liu2017video}, ToFlow~\cite{xue2019video} and CyclicGen~\cite{liu2019deep}, but they are unable to compete with SepConv~\cite{niklaus2017videosepcov} or CtxSyn~\cite{niklaus2018context}, as reported in~\cite{niklaus2020softmax}. In addition, the preliminary work~\cite{ding2021cdfi} triggers many recent researches in developing efficient models, such as DRVI~\cite{wu2021drvi} and $L^2BEC^2$~\cite{zhang2}, but their results are not comparable to ours. Lastly, we observe that AdaCoF~\cite{lee2020adacof} is only average among the other approaches, but our final model, which is built on AdaCoF, has arguably the best overall performance while maintaining compactness, demonstrating the superiority of the proposed sparsity-guided design framework.

\subsection{Qualitative Evaluation}

We present the visual comparisons on the DAVIS dataset~\cite{perazzi2016} in Figure~\ref{fig:visual}. The first and third example contain complex motion and occlusion, while the second example involves many non-stationary finer details. Note that AdaCoF+~\cite{lee2020adacof} generates relatively hazy interpolation frames for each of these cases (see the motorbike, house and swing stool). In contrast, thanks to our newly incorporated FP module and PS mechanism, the outcomes predicted by our method based on it are more precise and realistic. Additionally, we compare with BMBC~\cite{park2020bmbc}, CAIN~\cite{choi2020channel} and EDSC~\cite{cheng2020multiple}. EDSC utilizes deformable separable convolution but estimates an additional mask to aid in image synthesis. However, they are less appealing than our method on the provided examples. One can see that their interpolations typically contain apparent artifacts and are incapable of preserving clear features. Note that BMBC~\cite{park2020bmbc} occasionally produces sharp results but not as consistently as we do. We conjecture that the bilateral cost volume in BMBC improves the estimations of intermediate motion, which can also be incorporated into our design. Note that our model is the smallest among them, which once again demonstrates the benefit of the sparsity-guided network design.

\section{Conclusion}\label{sec:con}

We presented a sparsity-guided network design for frame interpolation that employs model compression as a guide for selecting an efficient architecture, which we then improved. As an instance, we showed that a considerably smaller AdaCoF model performs comparably to the original one, and with simple modifications it is able to significantly outperform the baseline and is also superior to other state-of-the-art methods. We emphasize that the optimization-based compression over a baseline model is independent of the baseline's specific architecture. Therefore, we believe that our framework is generic to be extended to various models and provides \emph{a new perspective} on the design of effective frame interpolation algorithms. In future work, it will be beneficial to establish a strong connection between the compression and design stages which iteratively improves the underlying architecture. 


%





\ifCLASSOPTIONcaptionsoff
  \newpage
\fi



\bibliographystyle{IEEEtran}
\bibliography{bare_adv}
\nocite{stergiou2021adapool,lee2022beyond,choi2021motion, liu2022jnmr, yang2022beyond}

%



%

\vspace{-.4in}

\begin{IEEEbiography}[{\includegraphics[width=1in,height=1.5in,clip,keepaspectratio]{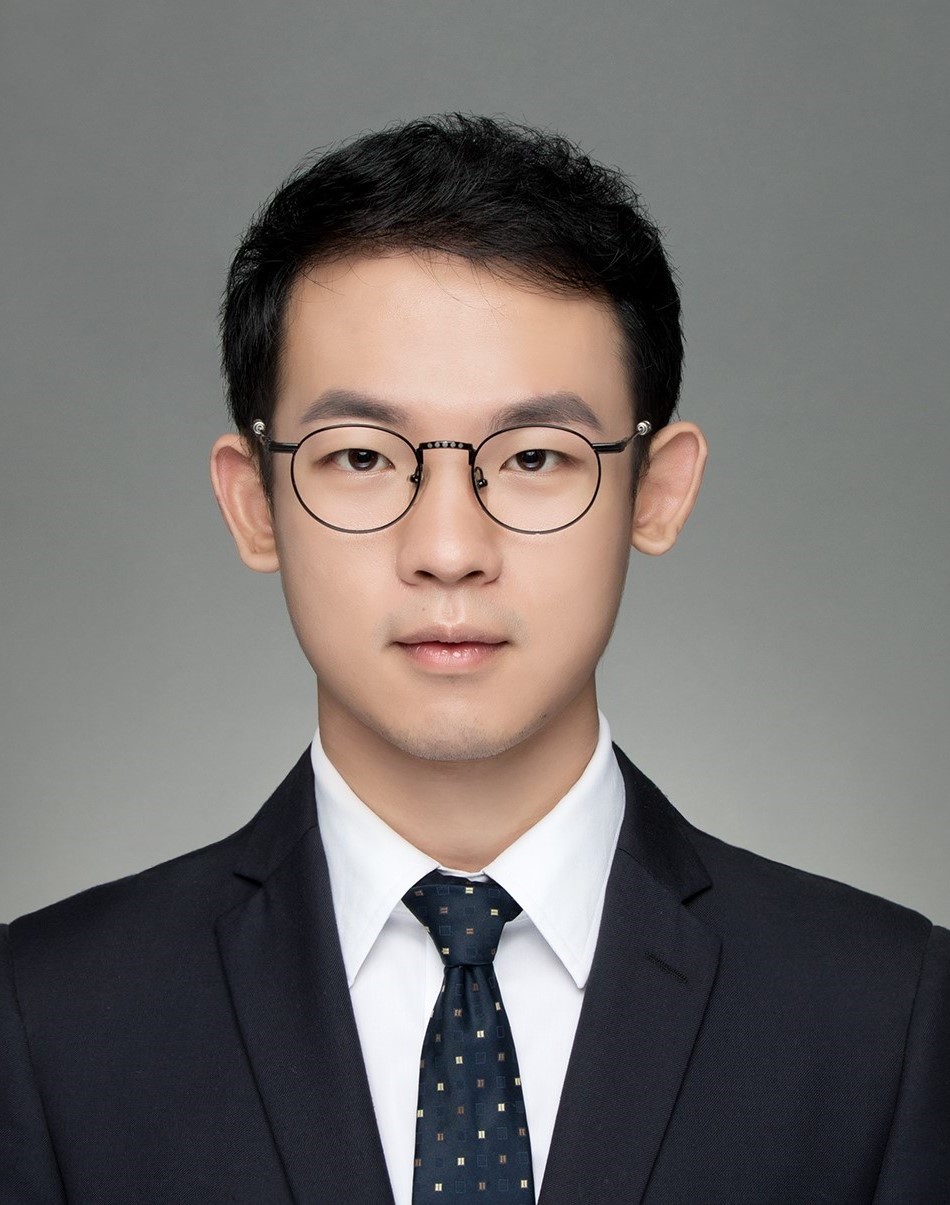}}]{Tianyu Ding} is currently a Senior Researcher at Microsoft, Redmond, USA. He received his PhD degree in Applied Mathematics and Statistics from Johns Hopkins University (JHU). Before that, he received two Master's degrees in Computer Science and Financial Mathematics from JHU. He received his Bachelor's degree in Mathematics from Sun Yat-sen University. His research interests are in computer vision, deep learning and numerical optimization.
\end{IEEEbiography}

\vspace{-.4in}

\begin{IEEEbiography}[{\includegraphics[width=1in,height=1.5in,clip,keepaspectratio]{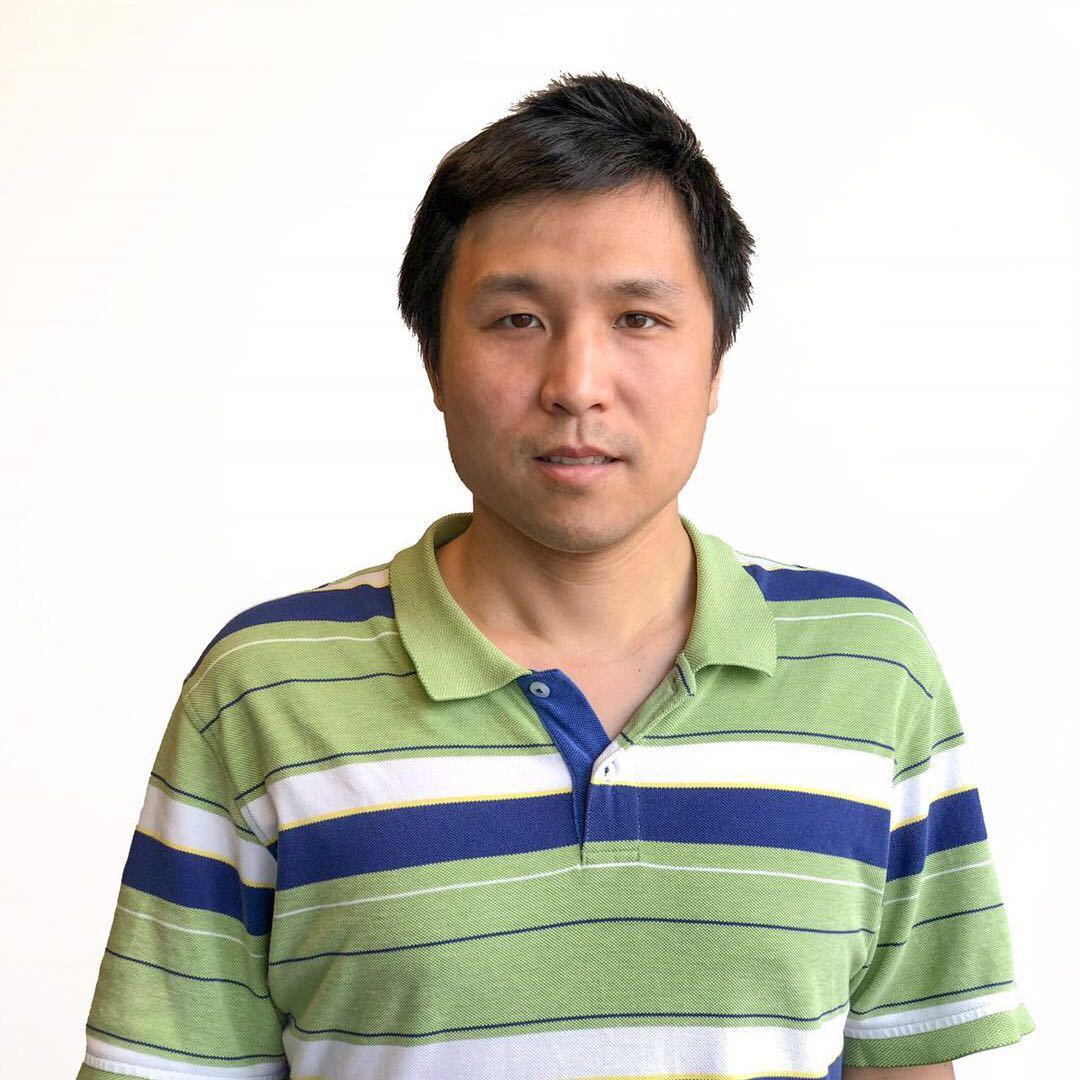}}]{Luming Liang BS05'; MEng08'; PhD14'} is currently a Principal Research Manager at Applied Sciences Group, Microsoft, Redmond, USA. He got his Ph.D. in 2014 from the Colorado School of Mines, majored in Computer Science, with a minor in Geophysics. His research interests are in computer vision, computer graphics, deep learning and signal processing. 
\end{IEEEbiography}

\vspace{-.4in}

\begin{IEEEbiography}[{\includegraphics[width=1.2in,height=1.3in,clip,keepaspectratio]{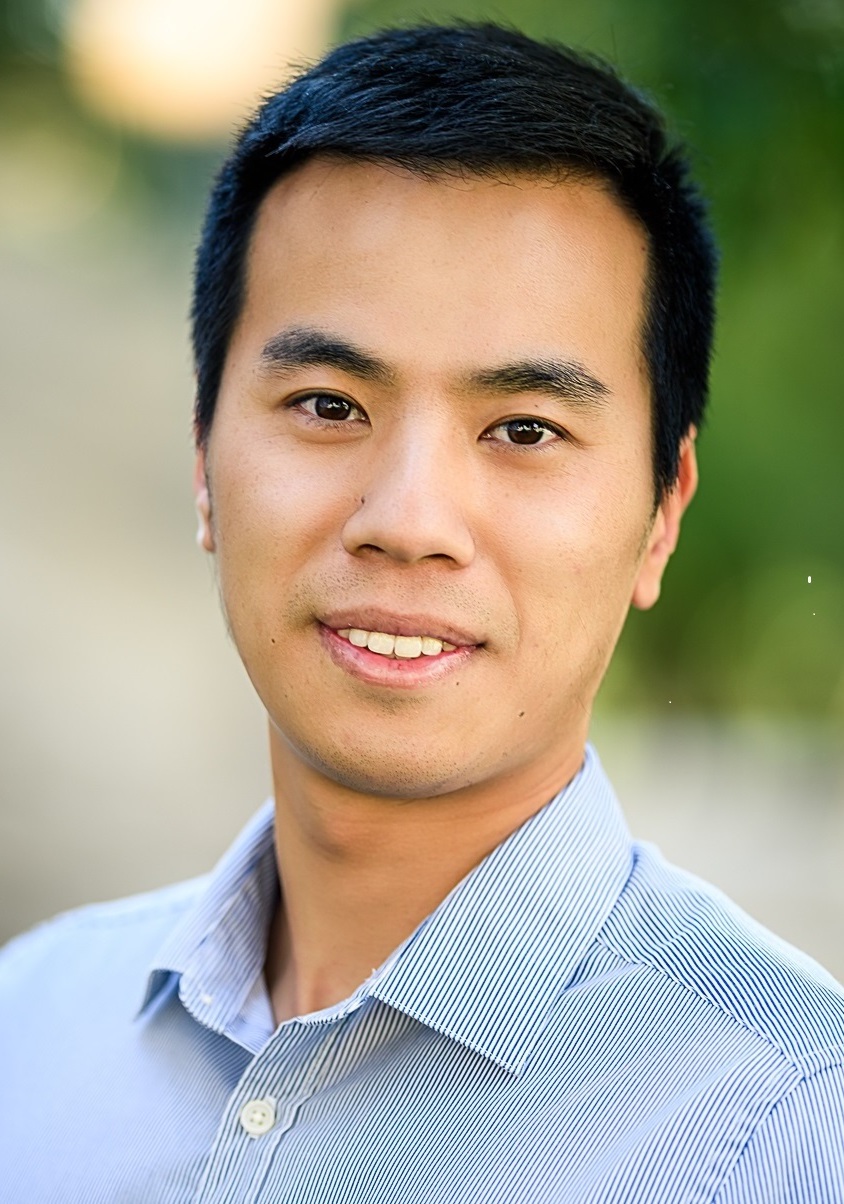}}]{Zhihui Zhu} is currently an Assistant Professor with the Department of Computer Science and Engineering at the Ohio State University. He was an Assistant Professor in the Department of Electrical and Computer Engineering at the University of Denver from 2020-2022 and a Post-Doctoral Fellow with the Mathematical Institute for Data Science, Johns Hopkins University, from 2018 to 2019. He received his Ph.D. degree in electrical engineering in 2017 from the Colorado School of Mines, where his research was awarded a Graduate Research Award. His research focuses on modeling and algorithmic aspects of data science and machine learning. He is or has been an Action Editor of the Transactions on Machine Learning Research and an Area Chair for NeurIPS. 
\end{IEEEbiography}

\vspace{-.4in}

\begin{IEEEbiography}[{\includegraphics[width=1in,height=1.25in,clip,keepaspectratio]{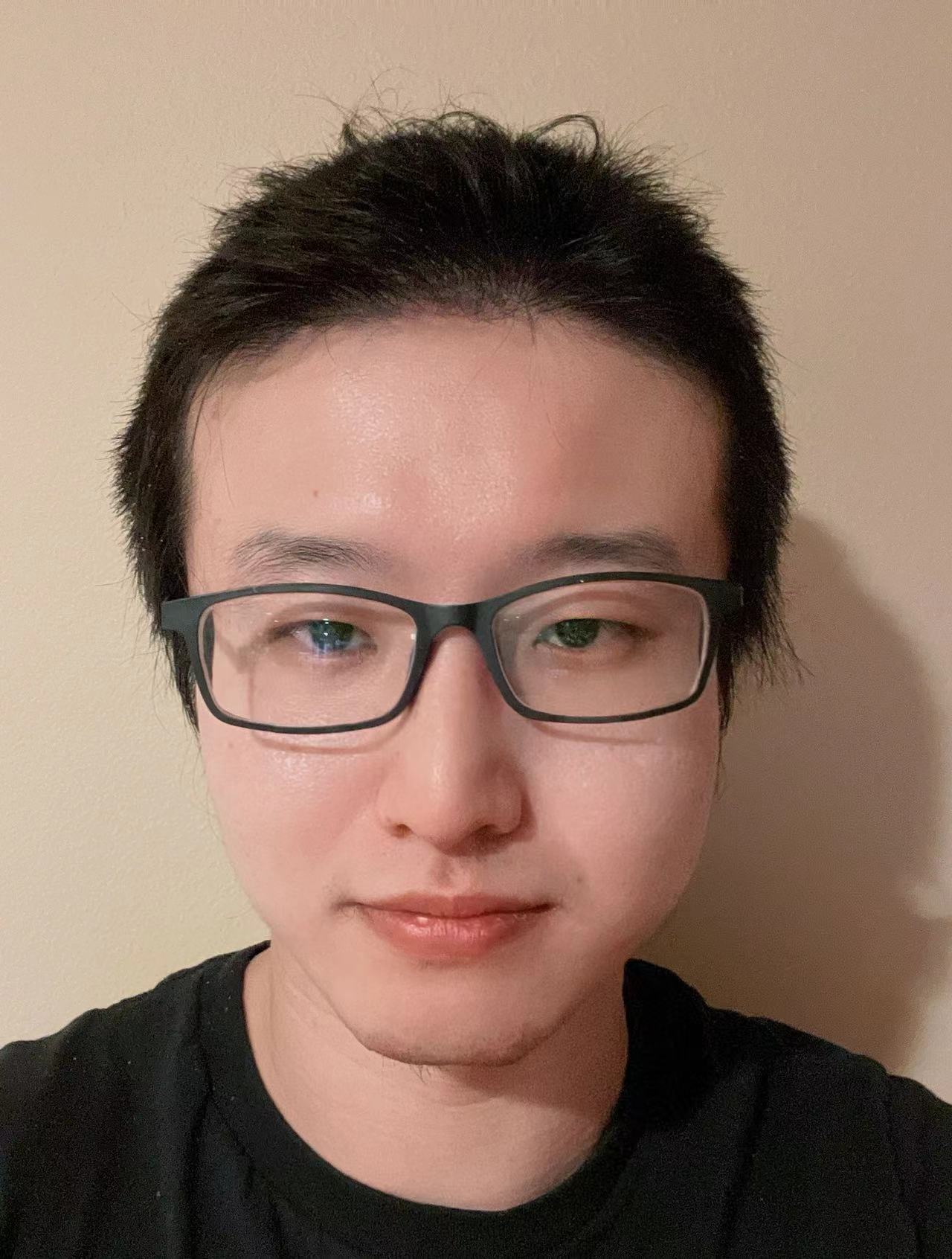}}]{Tianyi Chen}
Tianyi Chen is a Senior Researcher in Microsoft, Redmond, USA. He received his PhD degree in Applied Mathematics and Statistics and Master degree in Computer Science from Johns Hopkins University, and Bachelor degree in Mathematics from Dalian University of Technology. 
His research interests fall in numerical optimization and its applications in deep learning ranging from computer vision to natural language processing.  
\end{IEEEbiography}

\vspace{-.4in}

\begin{IEEEbiography}[{\includegraphics[width=1in,height=1.5in,clip,keepaspectratio]{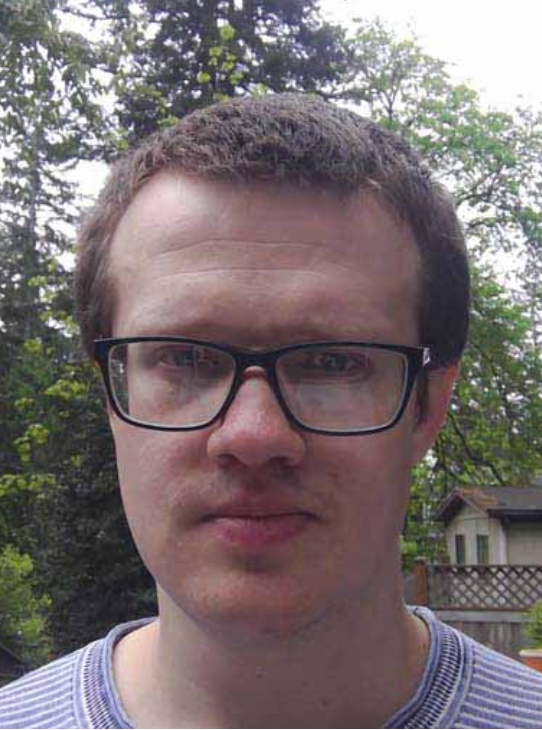}}]{Ilya Zharkov} is currently a Principal Research Manager at at Applied Sciences Group Microsoft, Redmond, USA. Before joining Microsoft in 2017, he worked on automated map generation from aerial imagery, optical character recognition (OCR), and handwriting text recognition.
Ilya graduated with an M.S. in physics from the Moscow State University in 2005.
His current research interests include people and object detection, segmentation, tracking, and 3D reconstruction.
\end{IEEEbiography}






\end{document}